\pdfoutput=1
\documentclass[letterpaper]{article}
\usepackage{proceed2e}
\usepackage[margin=1in]{geometry}

\usepackage{times}

\usepackage{graphicx}
\graphicspath{{images/}}

\usepackage{amsfonts} %
\usepackage{amsmath}  %
\usepackage{bm}  %

\usepackage{pifont}%
\usepackage{tikz}
\usetikzlibrary{bayesnet}
\usetikzlibrary{arrows.meta,arrows}
\usetikzlibrary{decorations.pathreplacing}
\usetikzlibrary{shapes}
\usetikzlibrary{patterns}

\usepackage{natbib2}
\renewcommand{\bibsection}{}  %

\usepackage[tight,bf]{subfigure}
\newlength{\figwidth}
\newlength{\figheight}
\newlength{\figheighttwo}

\usepackage{booktabs}
\usepackage{multirow}

\usepackage{rotating}

\usepackage{adjustbox}
\usepackage{array}

\usepackage{xspace}
\makeatletter
\DeclareRobustCommand\onedot{\futurelet\@let@token\@onedot}
\def\@onedot{\ifx\@let@token.\else.\null\fi\xspace}

\makeatother

\usepackage[colorinlistoftodos,prependcaption,textsize=tiny]{todonotes}

\usepackage{titlesec}
\titlespacing*{\paragraph}{0pt}{0.0ex plus .2ex}{1em}
\usepackage{multicol}
\usepackage{float}
\floatstyle{plain}
\newfloat{twocolequfloat}{b}{zzz}
\floatname{twocolequfloat}{Equation}

\usepackage[font=small,labelfont=sl,textfont=sl]{caption}

\usepackage{placeins}
\usepackage{flushend}

\usepackage[colorinlistoftodos,prependcaption]{todonotes}

\title{DP-GP-LVM: A Bayesian Non-Parametric Model for Learning\\Multivariate Dependency Structures}

 \author{ {\bf Andrew R.~Lawrence} \\
 Department of Computer Science \\
 University of Bath \\
 Bath, United Kingdom \\
 \And
 {\bf Carl Henrik Ek} \\
 Department of Computer Science \\
 University of Bristol \\
 Bristol, United Kingdom \\
 \And
 {\bf Neill D.\,F.~Campbell} \\
 Department of Computer Science \\
 University of Bath \\
 Bath, United Kingdom \\
 }

\begin{document}

\maketitle

\begin{abstract}
We present a non-parametric Bayesian latent variable model capable of learning  dependency structures across dimensions in a multivariate setting. Our approach is based on flexible Gaussian process priors for the generative mappings and interchangeable Dirichlet process priors to learn the structure. The introduction of the Dirichlet process as a specific structural prior allows our model to circumvent issues associated with previous Gaussian process latent variable models. Inference is performed by deriving an efficient variational bound on the marginal log-likelihood on the model.
\end{abstract}

\section{Introduction} \label{sec:intro}
Latent variable models provide data-efficient and interpretable descriptions of data. By specifying a generative model, it is possible to achieve a compact representation through exploiting dependency structures in the observed data. Their probabilistic structure allows the model to be integrated as a component in a larger system and facilitates tasks such as data-imputation and synthesis. 

Efficient representations can be achieved when the intrinsic dimensionality of the data is much lower than in its observed representation. Traditional approaches, such as probabilistic PCA~\citep{probabilistic_pca} and GP-LVM~\citep{gplvm} make the assumption that the data lies on a single low-dimensional manifold embedded in the high-dimensional space. 

However, in many scenarios, this assumption is too simplistic as more intricate dependency structures are present in the data. In specific, there are many situations where groups of dimensions co-vary. For a human walking, each limb shares a variation from the direction of travel, similarly one would expect that both arms share information that is not always present in the lower limbs, however, it is not inconceivable that the left-side limbs share information not present on the right. 

Variations that are not common across all observed dimensions are challenging to model. If included in the latent representation, a variation only present in a subset of the dimensions will ``pollute'' the representation of the dimensions that do not share this characteristic. 

One approach to circumvent this issue is to learn a factorised latent representation where independent latent variables describe each group of variations. An example of such approach is the Inter-Battery Factor Analysis Model~\citep{ibfa} where the latent space consists of dimensions encoding structure shared across all dimensions separately from structure that is private within a group of variates. 

This model, and the approaches building on this idea, assume that the grouping of the observed dimensions is know a priori. Importantly, this means that the learning task is to recover a latent representation that reflects a give grouping of the dimensions in the observed space. 
Even for familiar data, such as human motion, specifying these groupings is challenging while, in other tasks, extracting the groupings themselves is essential. We refer to these groupings as `views'. One such example is a medical scenario where each observed variate corresponds to a specific, potentially costly and for the patient intrusive, medical test. If we can learn the groupings of variations we can potentially reduce the range of tests needed for diagnosis.

In this paper, we describe a latent variable model, which we term the DP-GP-LVM, that automatically learns the grouping of the observed data thereby removing the need for a priori specification. By formulating the generative model non-parametrically, our approach has unbounded representative power and can infer its complexity from data. The Bayesian formulation enables us to average over all possible groupings of the observations allowing the structure to emerge naturally from the data. We perform approximate Bayesian inference by optimising a lower bound on the marginal likelihood of the model.

\section{Background} \label{sec:background}

\setlength{\figwidth}{0.32\textwidth}
\setlength{\figheight}{0.22\textwidth}
\setlength{\figheighttwo}{0.32\textwidth}

\tikzstyle{latent} = [circle, fill=white, draw=black, scale=0.75, minimum size=32pt]

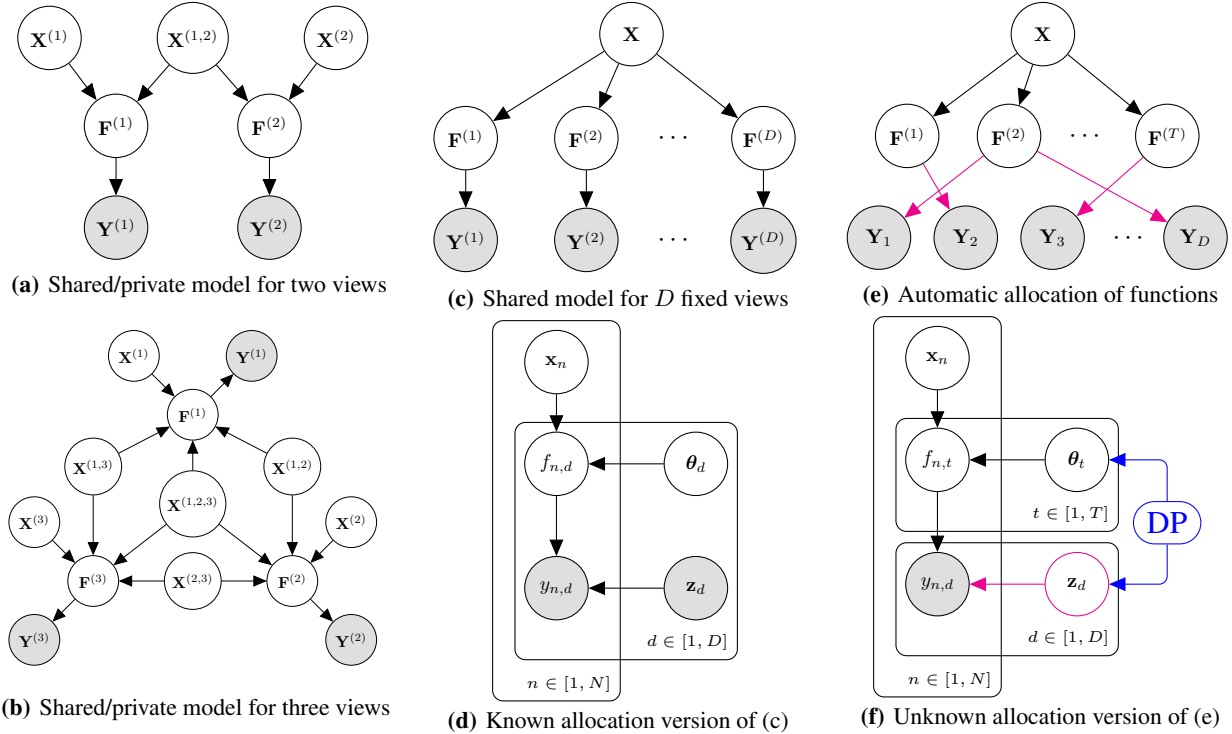
\begin{figure*}[tb]
  \centering
    \begin{minipage}[t]{\figwidth}\centering
    \subfigure[Shared/private model for two views]{%
      \centering%
      \begin{minipage}[t][\figheight]{\figwidth}\centering
        \begin{tikzpicture}

          \node[latent] (xs) {$\mathbf{X}^{(1,2)}$};
          \node[latent, left=1.0 of xs] (x1) {$\mathbf{X}^{(1)}$};
          \node[latent, right=1.0 of xs] (x2) {$\mathbf{X}^{(2)}$};
          \node[latent, below left=0.75 of xs, xshift=+0.2cm] (f1) {$\mathbf{F}^{(1)}$};
          \node[latent, below right=0.75 of xs, xshift=-0.2cm] (f2) {$\mathbf{F}^{(2)}$};
          \node[obs, below=0.5 of f1] (y1) {$\mathbf{Y}^{(1)}$};
          \node[obs, below=0.5 of f2] (y2) {$\mathbf{Y}^{(2)}$};

          \edge {xs} {f1, f2};
          \edge {x1} {f1};
          \edge {x2} {f2};
          \edge {f1} {y1};
          \edge {f2} {y2};

        \end{tikzpicture}
      \end{minipage}
        \label{fig:shared_two_views}
    }%
    \\[-5pt]%
    \subfigure[Shared/private model for three views]{%
      \centering%
      \begin{minipage}[c][\figheighttwo]{\figwidth}\centering
        \scalebox{0.8}{\begin{tikzpicture}

          \node[latent] (x123) {$\mathbf{X}^{(1,2,3)}$};
          \node[latent, above=0.5 of x123] (f1) {$\mathbf{F}^{(1)}$};
          \node[latent, below=0.25 of x123] (x23) {$\mathbf{X}^{(2,3)}$};
          \node[latent, left=0.75 of x23] (f3) {$\mathbf{F}^{(3)}$};
          \node[latent, right=0.75 of x23] (f2) {$\mathbf{F}^{(2)}$};
          \node[latent, above=1.0 of f2] (x12) {$\mathbf{X}^{(1,2)}$};
          \node[latent, above=1.0 of f3] (x13) {$\mathbf{X}^{(1,3)}$};
          \node[latent, above left=0.53 of f3] (x3) {$\mathbf{X}^{(3)}$};
          \node[obs, below left=0.53 of f3] (y3) {$\mathbf{Y}^{(3)}$};
          \node[latent, above right=0.53 of f2] (x2) {$\mathbf{X}^{(2)}$};
          \node[obs, below right=0.53 of f2] (y2) {$\mathbf{Y}^{(2)}$};
          \node[latent, above left=0.53 of f1] (x1) {$\mathbf{X}^{(1)}$};
          \node[obs, above right=0.53 of f1] (y1) {$\mathbf{Y}^{(1)}$};

          \edge {x123} {f1, f2, f3};
          \edge {x12} {f1, f2};
          \edge {x13} {f1, f3};
          \edge {x23} {f2, f3};
          \edge {x1} {f1};
          \edge {x2} {f2};
          \edge {x3} {f3};
          \edge {f1} {y1};
          \edge {f2} {y2};
          \edge {f3} {y3};

        \end{tikzpicture}}%
      \end{minipage}
        \label{fig:shared_three_views}
    }%
    \end{minipage}
    \hfill%
    \begin{minipage}[t]{\figwidth}\centering
    \subfigure[Shared model for $D$ fixed views]{%
      \centering%
      \begin{minipage}[t][\figheight]{\figwidth}\centering
        \begin{tikzpicture}

          \node[latent] (x) {$\mathbf{X}$};
          \node[latent, below=0.53 of x, xshift=-0.8cm] (f2) {$\mathbf{F}^{(2)}$};%
          \node[latent, left=0.75 of f2] (f1) {$\mathbf{F}^{(1)}$};
          \node[latent, right=1.5 of f2] (fv) {$\mathbf{F}^{(D)}$};%
          \node[obs, below=0.5 of f1] (y1) {$\mathbf{Y}^{(1)}$};
          \node[obs, below=0.5 of f2] (y2) {$\mathbf{Y}^{(2)}$};
          \node[obs, below=0.5 of fv] (yv) {$\mathbf{Y}^{(D)}$};

          \edge {x} {f1, f2, fv};
          \edge {f1} {y1};
          \edge {f2} {y2};
          \edge {fv} {yv};

          \node[right=0.4 of f2] {$\cdots$};%
          \node[right=0.4 of y2] {$\cdots$};%

        \end{tikzpicture}
      \end{minipage}
        \label{fig:shared_v_views}
    }%
    \\[-5pt]%
    \subfigure[Known allocation version of~\subref{fig:shared_v_views}]{%
      \centering%
      \begin{minipage}[c][\figheighttwo]{\figwidth}\centering
        \begin{tikzpicture}

          \node[latent] (x) {$\mathbf{x}_n$};
          \node[latent, below=0.5 of x] (f) {$f_{n,d}$};
          \node[latent, right=1.0 of f] (theta) {$\bm{\theta}_{d}$};
          \node[obs, below=0.8 of f] (y) {$y_{n,d}$};
          \node[obs, right=1.0 of y] (z) {$\mathbf{z}_d$};

          \edge {x, theta} {f};
          \edge {f, z} {y};

          \node[below=0.3 of y, xshift=+0.6cm] (dummy) {};
          \node[below=0.3 of y, xshift=-0.6cm] (dummy2) {};

          \plate {num_samples} {(x)(f)(y)(dummy)(dummy2)} {\tiny{$n\in[1,N]$}};
          \plate {num_views} {(f)(theta)(y)(z)} {\tiny{$d \in [1,D]$}};

        \end{tikzpicture}
      \end{minipage}
        \label{fig:shared_v_views_collapsed}
    }%
    \end{minipage}
    \hfill%
    \begin{minipage}[t]{\figwidth}\centering
    \subfigure[t][Automatic allocation of functions]{%
      \centering%
      \begin{minipage}[t][\figheight]{\figwidth}\centering
        \begin{tikzpicture}

          \node[latent] (x) {$\mathbf{X}$};
          \node[latent, below=0.5 of x, xshift=-0.6cm] (f2) {$\mathbf{F}^{(2)}$};
          \node[latent, left=0.5 of f2] (f1) {$\mathbf{F}^{(1)}$};
          \node[latent, right=1.25 of f2] (fv) {$\mathbf{F}^{(T)}$};
          \node[obs, below=0.5 of f1, xshift=-0.5cm] (y1) {$\mathbf{Y}_{1}$};
          \node[obs, right=0.3 of y1] (y2) {$\mathbf{Y}_{2}$};
          \node[obs, right=0.3 of y2] (y3) {$\mathbf{Y}_{3}$};
          \node[obs, below=0.5 of fv, xshift=+0.5cm] (yd) {$\mathbf{Y}_{D}$};

          \edge {x} {f1, f2, fv};
          \edge[magenta] {f1} {y2};
          \edge[magenta] {f2} {y1, yd};
          \edge[magenta] {fv} {y3};

          \node[right=0.25 of f2] {$\cdots$};
          \node[right=0.25 of y3] {$\cdots$};

        \end{tikzpicture}
      \end{minipage}
        \label{fig:unknown_views}
    }%
    \\[-5pt]%
    \subfigure[Unknown allocation version of~\subref{fig:unknown_views}]{%
      \centering%
      \begin{minipage}[c][\figheighttwo]{\figwidth}\centering
        \begin{tikzpicture}

          \node[latent] (x) {$\mathbf{x}_n$};
          \node[latent, below=0.5 of x] (f) {$f_{n,t}$};
          \node[latent, right=1.0 of f] (theta) {$\bm{\theta}_{t}$};
          \node[obs, below=0.8 of f] (y) {$y_{n,d}$};
          \node[latent, draw=magenta, right=1.0 of y] (z) {$\mathbf{z}_d$};

          \edge {x, theta} {f};
          \edge {f} {y};
          \edge[magenta] {z} {y};

          \node[below=0.3 of y, xshift=+0.6cm] (dummy) {};
          \node[below=0.3 of y, xshift=-0.6cm] (dummy2) {};

          \plate {num_samples} {(x)(f)(y)(dummy)(dummy2)} {\tiny{$n\in[1,N]$}};
          \plate {num_views} {(f)(theta)} {\tiny{$t \in [1,T]$}};
          \plate {num_dims} {(y)(z)} {\tiny{$d \in [1,D]$}};

          \node[draw, scale=1.2, rounded rectangle, below=0.12 of theta, blue, xshift=+1.0cm] (dp) {DP};
          \draw[rounded corners=3pt, ->, blue] (dp)|-(theta);
          \draw[rounded corners=3pt, ->, blue] (dp)|-(z);

        \end{tikzpicture}
      \end{minipage}
        \label{fig:unknown_views_collapsed}
    }%
    \end{minipage}\\[-10pt]%
    \caption{Graphical models for the Shared GP-LVM, MRD and our DP-GP-LVM (observed variables are shaded and latent variables unshaded). \subref{fig:shared_two_views}~The Shared GP-LVM allows for shared ($\mathbf{X}^{(1,2)}$) and private ($\mathbf{X}^{(1)}, \mathbf{X}^{(2)}$) latent variables. %
    \subref{fig:shared_three_views}~The addition of new observations leads to a combinatorial explosion in the number of latent variables.
    \subref{fig:shared_v_views}~The MRD model uses a single latent space with ARD parameters to allow sharing and   
    can be represented in collapsed form~\subref{fig:shared_v_views_collapsed} using a \textbf{known} assignment variable $\bf{Z}$ that allocates the ``views'' (groupings of data) to generative functions. %
    \subref{fig:unknown_views}~In an ideal model, we would infer the sharing of functions ({\color{magenta}{magenta}}) between observed dimensions from data. We would also like to infer $T$, the number of such functions, automatically from data.
     \subref{fig:unknown_views_collapsed}~In our DP-GP-LVM model we learn the \textbf{unknown} allocation $\bf{Z}$ into $T$ groupings (also unknown) automatically using a Dirichlet process prior ({\color{blue}{blue}}).} %
    \label{fig:shared_graphical_models}
\end{figure*}

Finding latent representations of data is a task central to many machine learning applications. By exploiting dependencies in the data, more efficient low dimensional representations can be recovered. Of specific importance is the work of~\cite{Spearman:1904tx} where the interpretation of a latent dimension as a \emph{factor} was introduced. The traditional factor analysis model is unidentifiable meaning additional assumptions need to be incorporated, for example, the Gaussian assumption made in PCA~\citep{pca}. 

A second approach is to introduce groupings of the observed data, as is done in CCA~\citep{Hotelling:1936wo}, where the latent representation that best describes the correlation between the groups is sought. While PCA was described as a model, it is challenging to describe the generative procedure of CCA. 

Another model that exploits groupings of the observed data is the Inter Battery Factor Analysis model (IBFA)~\citep{ibfa} where two different classes of latent factors were introduced. Specifically, the IBFA learns two separate factors, ``shared'' and ``private'', where the former represent variations common to all groups while the latter variations only belong to a single group. As discussed in the introduction, this factorisation is important when specifying a generative model. 

A Bayesian formulation of IBFA was proposed by~\cite{bayesian_cca} and a non-linear extension, based on Gaussian processes, called Manifold Relevance Determination (MRD) by~\cite{mrd}. However, there exists an important distinction between the two models that is rarely highlighted. The linear formulation of IBFA allows the groupings of the data to be inferred while, in the MRD model, the \emph{groupings} or \emph{views} needs to be set a priori. 

In this paper, we present a model that combines the benefit of both. We describe a non-linear, Bayesian model that allows the groupings to emerge from data. Our proposed model has significant commonalities with MRD and, being motivated by its shortcomings, we now proceed to describe MRD in detail.

\subsection{Latent Variable Models}
\label{sec:model}

A LVM aims to learn a latent representation $\mathbf{X}\in\mathbb{R}^{N\times Q}$ from a set of multivariate observations $\mathbf{Y}\in\mathbb{R}^{N\times D}$ where $N$ is the number of observations while $Q$ and $D$ are the dimensionality of the latent and observed data respectively. We denote the dimensions of the observed data as $\mathbf{y}_d$, $d\in[1,D]$ each consisting of $N$ observations $\mathbf{y}_d \in \mathbb{R}^{N}$. The generative model specifies the relationship between the latent space and the observed,
\begin{align}
  y_{n,d} = f_d(\mathbf{x}_n) + \epsilon_{n,d},
\end{align}
where the form of the noise $\bm{\epsilon}$ leads to the likelihood of the data.

The MRD is a member of a larger class of models called Gaussian Process Latent Variable Models~\citep{gplvm} (GP-LVM) where a Gaussian process prior~\citep{gp} is placed over the generating mapping $f(\cdot)$. Under the assumption of Gaussian noise, it is possible to  marginalize over the whole space of functions leading to a rich and expressive model. 

Due to the non-linearities of $f(\cdot)$, integration of the latent variables cannot be achieved in closed form. Inference of $\mathbf{X}$ can either be achieved through maximum-likelihood or via approximate integration by optimising a variational lower bound on the marginal likelihood~\citep{bayesian_gplvm}. \cite{shared_gplvm}~defined a shared GP-LVM where the observed data was grouped into two sets of views sharing a single set of latent variables. 
\citet{Ek:2008up} extended this further by introducing the idea of a factorisation with shared and private latent space from IBFA.

\paragraph{Views}
We use the term ``views'' to refer to a natural groupings within a set of data; data within a specific view will share a causal root or generative structure. 
Thus, views are observed data that are aligned in terms of samples but of a different modality or a disjoint group of observed dimensions. 
For example, we might consider a dataset consisting of silhouettes of people, the angles of the joints between their limbs and the background appearance of a room. These would consist of three views (silhouettes, angles and appearance) where the silhouettes and angles have shared (the pose of the body) and private (clothes affect the silhouette but not the joints) information. The appearance of the background is a third view that should be independent of the first two since it has no causal link.

\paragraph{Model Evolution} We now describe the sequence of graphical models in Figure~\ref{fig:shared_graphical_models} from the model described by \cite{Ek:2008up} in \subref{fig:shared_two_views} to our proposed model in \subref{fig:unknown_views_collapsed}. The model by \cite{Ek:2008up} naturally extends beyond two separate groupings. However, due to the fixed structure of the latent space, it leads to a combinatorial explosion in the number of latent variables illustrated by Figure~\ref{fig:shared_three_views}. Further, learning is challenging as the dimensionality of each latent space needs to be known a priori. 

To circumvent these issues, the MRD model, Figure~\ref{fig:shared_v_views}, treats the factorisation as part of the GP prior. This GP prior is completely specified by its mean and covariance function. For most unsupervised tasks an zero mean function is assumed, leaving only the covariance function as its parametrisation. The introduction of Automatic Relevance Determination (ARD) \citep{Neal1996} covariance functions allows the MRD to enclose the factorisation into the GP prior.

\paragraph{ARD} In a stationary kernel, the covariance between two latent variables is a function of the distance between the points. Rather than a spherical distance function, the ARD version introduces a parametrised diagonal Mahalanobis distance that is learned independently for each view (grouped observations). The intuition is now that if the distance function ``switches off'' an axis, this view becomes independent of the corresponding latent dimension and the factorization can be determined by the non-zero ARD weights. 

However, this approach leads to additional problems as the ARD parameters can also be interpreted as an inverse length scale. This means that a small ARD value for a specific dimension could have two different causes; either that the dimension vary linearly with the view or because it is invariant to the view~\cite{aki_thesis}. While the MRD only considers the latter cause, we explicitly model both these cases; this is our first extension to the MRD.

\paragraph{Inference of Views (Groupings)} In the MRD model, the groupings of the observed variates must be specified a priori. This (i)~restricts the data that can be used, (ii) can be very challenging to specify without supervision, and (iii) the representation will be sensitive to changes or errors in the grouping provided. 

One approach to circumvent this is outlined by~\cite{mrd2}, shown in Figure~\ref{fig:shared_v_views_collapsed}, where a separate function is used for each dimension and then a clustering is performed as a post-processing step. In addition to the unsatisfactory post-processing, this solution will lead to a significant increase in the number of parameters as a Mahalanobis metric needs to be learned for each output dimension. 

In this paper, we introduce a specific unknown indicator variable $\mathbf{z}_d$ that determines which latent dimensions will be associated with each output dimension as in Figure~\ref{fig:unknown_views_collapsed}. Further, in order to control the structure of the latent space, we introduce a Dirichlet Process (DP) prior that allows us to include prior knowledge of the complexity of the latent representation.

\paragraph{Other Models using Stochastic Processes}
Previous models have combined elements of GPs with DPs. %
Mixtures of GP experts place a Gaussian mixture model on the input space then fit each GP to the data belonging to the specific components.

An infinite mixture of GP experts uses a DP to determine the number of components. The main works using this approach are by~\cite{NIPS2001_2055} and~\cite{NIPS2005_2768}, who use MCMC to approximate the intractable posterior, as well as~\cite{NIPS2008_3395} and~\cite{5664792}, who use variational inference.

In addition,~\cite{hensman2015fast} combines a DP and GP for the purpose of clustering time-series data streams. However, as with the mixture of GP experts, their model focuses on supervised learning and does not address the unsupervised task that we are studying.

The topic is related to work by \cite{palla_nips_12} on variable clustering using DPs to infer block-diagonal covariance structures in data. \cite{wood_uai_06} defined a prior over a number of hidden causes and used reversible jump Markov Chain Monte Carlo to approximate a distribution over causal structures in data.

A related avenue of investigation is the multi-output GP literature, for example~\cite{NIPS2008_3553,pmlr-v9-alvarez10a,multiout_gp_jmlr,NIPS2017_7098}. These works, looking at transfer learning or filling in missing data, also produce structured models and a number have made use of the Indian Buffet Process (in contrast to a DP) to control complexity and favour sparse explanations in these models.

\section{The DP-GP-LVM Model} \label{sec:gp-dp}

\newcommand{\Kxx}{\mathbf{K}_{\text{xx}}}
\newcommand{\Kd}{\mathbf{K}_{\bm{\theta}_d}}
\newcommand{\Kt}{\mathbf{K}_{\bm{\theta}_t}}

\paragraph{Generative Model} We assume that the observed data are generated as a function of some unknown latent variables $\mathbf{X} \in \mathbb{R}^{N \times Q}$ where the $N$ observations each come from a lower $Q$ dimensional latent space such that $Q \ll D$. Thus we may write
\begin{equation}
\mathbf{y}_d = f_{d}( \mathbf{X}  ) + \bm{\epsilon}_d \label{eqn:basic_unsupervised}
\end{equation}
where $f_{d}(\cdot)$ is some function and $\bm{\epsilon}_d \sim \mathcal{N}(\mathbf{0},\beta_d^{-1}\mathbf{I}_N)$ is zero mean iid Gaussian noise with precision $\beta_d$. We put a standard Gaussian prior over the latent space,
\begin{equation}
p(\mathbf{X}) = \prod_{q=1}^{Q} \mathcal{N}\!\left( \mathbf{x}_q \vert \bm{0}, \mathbf{I}_{N} \right), \label{eqn:model_factor_x}
\end{equation}
and place a zero mean Gaussian process prior over the function such that
\begin{align}
  \mathbf{f}_d(\mathbf{X}) &\sim \mathcal{GP}\!\left(\mathbf{0}, k(\mathbf{X},\mathbf{X}^{\prime} \vert \bm{\theta}_d) \right) \label{eqn:gp_prior}\\
  p(\mathbf{F} \vert \mathbf{X}) &= \prod_{d=1}^{D}{\mathcal{N}(\mathbf{f}_d ; \mathbf{0},\Kd)},
\end{align}
where $\mathbf{f}_d \in \mathbb{R}^{N}$ denotes the evaluation of the function at the latent locations $X$ and $\Kd = k(\mathbf{X},\mathbf{X}^{\prime} ; \bm{\theta}_d)$ denotes the evaluation of some covariance function $k(\cdot,\cdot)$ with hyperparameters $\bm{\theta}_d$. As for the other random variables, we use $\mathbf{F}$ to denote the concatenation of $\mathbf{f}_d$ across $d$.

The observed data is then obtained from these latent functions through a likelihood to model the Gaussian noise
\begin{equation}
p(\mathbf{Y} \vert \mathbf{F}) = \prod_{d=1}^{D}{\mathcal{N}(\mathbf{y}_d \vert \mathbf{f}_d, \beta_d^{-1}\mathbf{I}_N)}.
\end{equation}

\paragraph{Sharing Functions} As we have discussed previously, %
we assume that our multivariate observations are not all independent but will potentially share generative correlations. %
From this assumption, there are two properties we would like to encode in our model. Firstly, we would like to encourage the observations to be grouped together, when the data supports it, and share a common generative function
\begin{equation}
\mathbf{y}_{d'} = f_{t}( \mathbf{X}  ) + \bm{\epsilon}_{t} \quad \forall \quad d' \in \mathcal{D}_t, \label{eqn:grouped_functions}
\end{equation}
where $\mathcal{D}_t$, $t \in [1,\infty]$ denotes a grouped subset of observed dimensions such that
$\bigcup_{t=1}^{\infty} \mathcal{D}_t = [1,D]$.

Secondly, we do not know a priori what these groupings should be and therefore $\{\mathcal{D}_t\}$ must be inferred from the data itself. In general, there could be an infinite set of potential groupings, however we note that in practice $\vert\{\mathcal{D}_t\}\vert \leq D$.
We now describe how we achieve the sharing of functions and the inference over groupings, a key contribution of our approach.

\paragraph{Function Parameterization} The differences in the shared $f_{t}(\cdot)$ functions in~\eqref{eqn:grouped_functions} are encoded by the hyperparameters of the covariance functions in the GP prior~\eqref{eqn:gp_prior}. We can adopt a covariance function that makes use of Automatic Relevance Determination (ARD) to infer a subset of the $Q$ latent dimensions to be used. We make use of a squared exponential covariance function
\begin{equation}
k(\mathbf{x}_{i},\mathbf{x}_{j}; \bm{\theta}_t) = {\sigma_t}^2 \exp{\!\left( -\frac{1}{2} \sum_{q=1}^{Q}{\gamma_{t,q} (x_{i,q}-x_{j,q})^{2}} \right)}\,,
\end{equation}
where the hyperparameters $\bm{\theta}_t = [\sigma_t^{2}, \bm{\gamma}_t]$ are the signal variance $\sigma_t^{2}$ and the positive ARD weights $\bm{\gamma}_{t} \in \mathbb{R}_{+}^Q$. We observe that if $\gamma_{t,q'} \rightarrow 0$ then the function has no dependence on the $q'$ dimension of the latent space; the function is independent of this latent dimension.

\paragraph{Grouping Assignments} We specify the assignment of the observed dimensions $d$ to the appropriate groups $\{\mathcal{D}_t\}$ using a multinomial assignment variable $\mathbf{z}_d$ such that
\begin{align}
p(\mathbf{f}_d \vert \mathbf{X}, \mathbf{z}_d, \bm{\Theta}) &= \prod_{t=1}^{\infty} {\mathcal{N}(\mathbf{f}_d \vert \mathbf{0},\Kt^{[\mathbf{z}_d = t]})} \label{eqn:model_factor_f}\\
p(\mathbf{y}_d \vert \mathbf{f}_d, \mathbf{z}_d, \bm{\beta}) &= \prod_{t=1}^{\infty}{\mathcal{N}(\mathbf{y}_d \vert \mathbf{f}_d, \beta_t^{-[\mathbf{z}_d = t]}\mathbf{I}_N)}, \label{eqn:model_factor_y}
\end{align}
where $\bm{\Theta} = \{\bm{\theta}_t\}$, $\bm{\beta} = \{\beta_t\}$ and $[\cdot]$ is the Iverson bracket notation for the indicator function.

\paragraph{Dirichlet Process Prior} Since the grouping dependence in~\eqref{eqn:grouped_functions} is encoded in the hyperparameters, we can encode our preference for sharing and the inference over the groupings by placing a Dirichlet Process (DP) prior over the hyperparameters (and noise precision) of the covariance functions for each observed dimension $d$. The DP consists of a base measure and a clustering parameter $\alpha$. We use a wide log-Normal distribution as the base measure. By drawing the hyperparameters $\bm{\theta}_d$ from this log-Normal prior via a DP,
\begin{align}
\bm{\Theta}, \bm{\beta} &\sim  \mathcal{DP}\!\left(\alpha, p(\bm{\Theta}, \bm{\beta} ) \right)  \label{eqn:dp_prior} \\
p(\bm{\Theta}, \bm{\beta} ) &= \log\mathcal{N}(\bm{0}, \mathbf{I}_{Q+1}) \label{eqn:model_factor_theta},
\end{align}
the hyperparameters will be clustered; all the output dimensions sharing the same set of hyperparameters are effectively combined to form the %
set of groupings $\{\mathcal{D}_t\}$.

\paragraph{Stick Breaking Construction} To obtain the multinomial assignment variable $\mathbf{Z}$, we use the stick breaking construction of a DP. We obtain a, potentially infinite, set of stick lengths $v_t \in [0,1]$ through independent draws from a Beta distribution
\begin{equation}
p(v_t \vert \alpha) \sim  \text{Beta}(1, \alpha)  \label{eqn:model_factor_v}%
\end{equation}
using the clustering parameter $\alpha$. From these stick lengths, we obtain a vector of mixing proportions
\begin{equation}
\pi_t(\mathbf{V}) = v_t \prod_{i=1}^{t-1} (1 - v_i).
\end{equation}
We use these to define the iid Multinomial distribution over the assignment variable
\begin{equation}
p(\mathbf{z}_d \vert \mathbf{V}) \sim \text{Mult}\!\left( \pi(\mathbf{V}) \right), \label{eqn:model_factor_z}
\end{equation}
where we may consider $\mathbf{z}_d$ as a one-hot encoding vector of dimension $d$ belonging to set $\mathcal{D}_t$.

\begin{figure}[tb]
  \centering
  \scalebox{0.85}{%
    \begin{tikzpicture}

      \node[latent] (alpha) {$\alpha$};
      \factor[below=0.5 of alpha] {alpha-prior} {right:$p(\alpha)$} {} {alpha};

      \node[latent, right=2.0 of alpha] (v) {$\mathbf{V}$};
      \factor[left=1.0 of v] {beta} {above:Beta$(1,\alpha)$} {alpha} {v};

      \node[latent, below=0.5 of v] (gamma) {$\bm{\gamma}$};
      \factor[left=0.75 of gamma, yshift=-0.5cm] {base-dist} {left:$p(\bm{\gamma})$} {} {gamma};

      \node[latent, below=1.0 of gamma] (sig-var) {$\sigma^{2}$};
      \factor[left=0.75 of sig-var] {sig-var-prior} {left:$p(\sigma^{2})$} {} {sig-var};

      \node[latent, below=0.5 of sig-var] (beta) {$\bm{\beta}$};
      \factor[left=0.75 of beta] {noise-prior} {left:$p(\bm{\beta})$} {} {beta};

      \node[latent, right=2.0 of v] (z) {$\mathbf{Z}$};
      \factor[right=0.8 of v] {multi} {above:Multi} {v} {z};

      \node[latent, below=0.5 of z] (f) {$\mathbf{F}$};

      \node[latent, right=1.0 of f] (x) {$\mathbf{X}$};
      \factor[above=0.5 of x] {x-prior} {above:$p(\mathbf{X})$} {} {x};

      \node[obs, right=2.0 of beta] (y) {$\mathbf{Y}$};

      \edge {gamma,sig-var,z,x} {f} ; %
      \edge {beta,f} {y} ; %

      \draw[bend right, ->] (z) to node [auto] {} (y);

      \plate {gamma_q} {(gamma)} {$Q$};

      \node[right=0.0 of beta, yshift=-0.5cm] (dummy) {};
      \plate [inner sep=0.2cm] {dp} {(v)(gamma_q)(beta)(sig-var)(dummy)} {$T \rightarrow \infty$};

      \node[ left=0.0 of y, xshift=+0.0cm] (dummy3) {};
      \node[right=0.0 of y, xshift=-1.3cm, yshift=-0.5cm] (dummy4) {};

      \plate [inner sep=0.2cm] {out_dims} {(y)(z)(dummy3)(dummy4)} {$D$};
      \plate {x_q} {(x)} {$Q$};

      \node[ left=0.0 of y, xshift=-0.0cm] (dummy2) {};
      \tikzset{plate caption/.append style={above left=0pt and 0pt of #1.south east}}
      \plate [inner sep=0.2cm] {num} {(y)(f)(x_q)(dummy2)} {$N$};

    \end{tikzpicture}}\\[-5pt]%
    \caption{Graphical model of DP-GP-LVM. The grey node $\mathbf{Y}$ is observed while all the white nodes represent latent random variables. During inference, we marginalize out all the latent variables except the hyperparameters and noise precision $\{\bm{\gamma}, \sigma_f^{2}, \bm{\beta}\}$ where we take MAP estimates.}%
    \label{fig:gp-dp-model}
\end{figure}
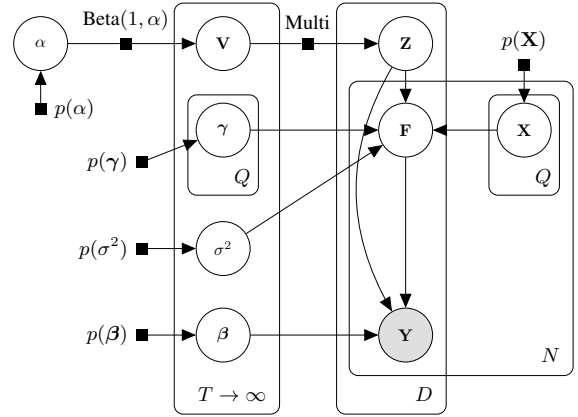

\paragraph{The Full Model} We combine all these terms to produce the full graphical model of Figure~\ref{fig:gp-dp-model}. The full joint distribution factorizes as
\begin{multline}
  p(\mathbf{Y}, \mathbf{F}, \mathbf{X}, \mathbf{Z}, \mathbf{V} \vert \alpha, \mathbf{\Theta}, \bm{\beta}) = p(\mathbf{Y} \vert \mathbf{F}, \mathbf{Z}, \bm{\beta}) p(\mathbf{Z} \vert \mathbf{V})   \\ 
  p(\mathbf{F} \vert \mathbf{X}, \mathbf{Z}, \mathbf{\Theta}) p(\mathbf{V} \vert \alpha) p(\bm{\Theta}, \bm{\beta}) p(\mathbf{X}) p(\alpha), \label{eqn:full_model}
\end{multline}
where the individual factors have been defined in \eqref{eqn:model_factor_y}, \eqref{eqn:model_factor_z}, \eqref{eqn:model_factor_f}, \eqref{eqn:model_factor_v}, \eqref{eqn:model_factor_theta}, \eqref{eqn:model_factor_x} and we use a wide Gamma prior over the clustering parameter
\begin{equation}
p(\alpha) \sim \text{Gamma}(\alpha \vert s_1, s_2). \label{eqn:model_factor_alpha}
\end{equation}
We perform learning by marginalizing out the latent variables $\{\mathbf{F}, \mathbf{X}, \mathbf{Z}, \mathbf{V}, \alpha\}$ and taking MAP estimates over the hyperparameters and noise precisions $\{\bm{\gamma}, \sigma_f^{2}, \bm{\beta}\}$. To deal with the intractable marginalizations, we use variational inference as will be described in \S~\ref{sec:gp-dp_learning}.

\subsection{Special Cases of DP-GP-LVM} The DP-GP-LVM model can be seen as a generalization of both the Bayesian GP-LVM~\citep{bayesian_gplvm} and the MRD~\citep{mrd} models.
We show this with reference to the full model of~\eqref{eqn:full_model} and Figure~\ref{fig:gp-dp-model}. 

\paragraph{Bayesian GP-LVM} In the Bayesian GP-LVM, the observed dimensions are assumed to be iid draws from the same function. This is captured in our model by taking the limiting case of a single cluster from the DP (such that $T \rightarrow 1$). In this setting, we have a single set of hyperparameters (and noise precision) shared across all dimensions $d$. This also means that latent variables $\{\mathbf{Z}, \mathbf{V}, \alpha\}$ may be removed from the model.

\paragraph{MRD} In the case of the MRD model, illustrated in Figures~\ref{fig:shared_v_views}~and~\ref{fig:shared_v_views_collapsed}, the grouping structure is specified a priori and not inferred from the data. In this instance, the allocation variable $\mathbf{Z}$ becomes observed (dictating the known allocation of dimensions into a finite set of $T$ groups $\{\mathcal{D}_t\}, t = [1,T]$) and the model collapses to that of MRD. The observation of $\mathbf{Z}$ then renders the variables $\{\mathbf{V}, \alpha\}$ unnecessary.

\newcommand{\margLogLike}{\log p(\mathbf{Y} \vert \bm{\Theta}, \bm{\beta})}
\newcommand{\Xu}{\mathbf{X}_{\text{u}}}
\newcommand{\bU}{\mathbf{U}}
\newcommand{\bF}{\mathbf{F}}
\newcommand{\bX}{\mathbf{X}}
\newcommand{\bY}{\mathbf{Y}}

\newcommand{\bV}{\mathbf{V}}
\newcommand{\bZ}{\mathbf{Z}}

\newcommand{\Kff}{\tilde{\mathbf{K}}_{\text{ff}}^{(d)}}
\newcommand{\Kuf}{\tilde{\mathbf{K}}_{\text{uf}}^{(d)}}
\newcommand{\Kfu}{\tilde{\mathbf{K}}_{\text{fu}}^{(d)}}
\newcommand{\Kuu}{\tilde{\mathbf{K}}_{\text{uu}}^{(d)}}

\newcommand{\betat}{\tilde{\beta}_d}

\newcommand{\variationalParams}{\bm{\mu}, \bm{\Sigma}, \Xu, \bm{a}, \bm{b}, \bm{\Phi}, \bm{w}}

\subsection{Learning} \label{sec:gp-dp_learning}

To perform learning of the joint model of~\eqref{eqn:full_model} we would like to marginalize out the latent variables $\{\mathbf{F}, \mathbf{X}, \mathbf{Z}, \mathbf{V}, \alpha\}$ and take MAP estimates over the hyperparameters and noise precisions $\{\bm{\Theta}, \bm{\beta}\}$. 
This corresponds to optimising the marginal log-likelihood of the observed data $\margLogLike$ which is
\begin{multline}
\log \hspace{-1em} \int\displaylimits_{\mathbf{F}, \mathbf{Z}, \mathbf{X}, \mathbf{V}, \alpha} \hspace{-1em} p(\mathbf{Y} \vert \mathbf{F}, \mathbf{Z}, \bm{\beta}) p(\mathbf{Z} \vert \mathbf{V}) p(\mathbf{V} \vert \alpha) \\[-10pt]
  p(\mathbf{F} \vert \mathbf{X}, \mathbf{Z}, \mathbf{\Theta})  p(\bm{\Theta}, \bm{\beta}) p(\mathbf{X}) p(\alpha). \label{eqn:marg_log_like}
\end{multline}
Unfortunately, a number of these integrals are intractable and cannot be found in closed form. To make progress, we introduce variational distributions to approximate the posteriors over the latent parameters and then optimize the Evidence Lower Bound (ELBO) in a similar manner to ~\cite{bayesian_gplvm} and~\cite{dp}.

\paragraph{Lower Bound} We introduce a factorized variational distribution $q(\mathbf{F}, \mathbf{X}, \mathbf{Z}, \mathbf{V}, \alpha)$ over the latent variables. 
If we define $\mathbf{\Omega} = \{\mathbf{F}, \mathbf{X}, \mathbf{Z}, \mathbf{V}, \alpha\}$ as the set of latent variables, we have 
\begin{align}
  \margLogLike &= \log{\int_{\mathbf{\Omega}} {q(\mathbf{\Omega}) \frac{p(\mathbf{Y},\mathbf{\Omega})}{q(\mathbf{\Omega})} }} \\
  &= \log{\left( \mathbb{E}_q \left[\frac{p(\mathbf{Y},\mathbf{\Omega})}{q(\mathbf{\Omega})}\right] \right)} \\
  &\geq \mathbb{E}_q \left[\log{p(\mathbf{Y},\mathbf{\Omega})}\right] + \mathbb{H}[q(\mathbf{\Omega})] , \label{eq:jensens}
\end{align}
with the lower bound $\mathcal{L} = \mathbb{E}_q \left[\log{p(\mathbf{Y},\mathbf{\Omega})}\right] + \mathbb{H}[q(\mathbf{\Omega})]$.
We decompose the lower bound into expressions from the GP ($\{\mathbf{F}, \mathbf{X}\}$) and the DP ($\{\mathbf{Z}, \mathbf{V}, \alpha\}$) such that
\begin{equation}
  \mathcal{L} = \mathcal{L}_{\mathcal{GP}} + \mathcal{L}_{\mathcal{DP}} + \log p(\bm{\Theta}, \bm{\beta}). \label{eqn:objective}
\end{equation}
We now describe each part of the lower bound.

\paragraph{GP Approximating Distributions} As noted by~\cite{bayesian_gplvm}, the lower bound on the GP
\begin{equation}
\small \log \! \int_{\bF,\bX} 
\! q(\mathbf{F})q(\mathbf{X}) \frac{p(\mathbf{Y} \vert \mathbf{F}, \mathbf{Z}, \bm{\beta}) p(\mathbf{F} \vert \mathbf{X}, \mathbf{Z}, \mathbf{\Theta}) p(\mathbf{X}) }{q(\mathbf{F})q(\mathbf{X})} %
\end{equation}
is still intractable due to the presence of $\mathbf{X}$ inside the covariance function in $p(\mathbf{F} \vert \mathbf{X}, \mathbf{Z}, \mathbf{\Theta})$. 
We make progress by extending the output space of the GP with a random variable $\mathbf{U}$ drawn from the same GP at some \emph{pseudo input locations} $\Xu$ such that $\mathbf{u}_d \sim \mathcal{GP}(0, k_f(\Xu, \Xu \vert \bm{\theta}_d)$.
These locations are taken as variational parameters and are optimized over through the lower bound.

If we assume that the $\bU$ form a sufficient statistic for the outputs $\bF$, then we have $p(\bF, \bU \vert \bX, \Xu) = p(\bF \vert \bU, \bX) p(\bU \vert \Xu)$. Further, if we assume that the approximating distribution factorises as $q(\bF,\bU,\bX) = p(\bF \vert \bU, \bX) q(\bU) q(\bX)$ then we have
\begin{equation}
\small \log \hspace{-0.6em} \int\displaylimits_{\bF,\bU,\bX} \hspace{-0.6em} p(\bF \vert \bU, \bX) q(\bU) q(\bX) \frac{p(\mathbf{Y} \vert \mathbf{F}, \mathbf{Z}, \bm{\beta}) p(\bU \vert \Xu) p(\mathbf{X}) }{q(\bU)q(\mathbf{X})}, 
\end{equation}
where all the terms are tractable. The optimal form of $q(\bU)$ is found to be Gaussian through variation calculus as shown by~\cite{bayesian_gplvm}; this distribution can then be marginalised out in closed form. A fully factorized Gaussian form is taken for $q(\bX)$ as
\begin{equation}
  q(\mathbf{X}) \sim \prod_{q=1}^{Q} \mathcal{N}(\mathbf{x}_{q} \vert \bm{\mu}_{q}, \bm{\Sigma}_{q}),
\end{equation}
where $\bm{\Sigma}_{q}$ are assumed to be diagonal.

\paragraph{GP Bound} This leads to a GP lower bound of 
\begin{equation}
  \mathcal{L}_{\mathcal{GP}} = \sum_{d=1}^{D}{\mathcal{F}_d} - \mathrm{KL}\!\left(q(\mathbf{X}) \!\parallel\! p(\mathbf{X})\right) \label{eqn:fd_kl_x}
\end{equation}
where the free energy $\mathcal{F}_d$ is given by
\begin{align}
 \mathcal{F}_d &= \frac{N}{2}\log{\betat} + \frac{1}{2}\log{\begin{vmatrix}\Kuu\end{vmatrix}} + \frac{\betat}{2}\mathrm{Tr}([\Kuu]^{-1}\bm{\Psi}_{2}) \nonumber \\
  &- \frac{N}{2}\log{(2\pi)} - \frac{1}{2}\log{\begin{vmatrix}\betat\bm{\Psi}_{2}+\Kuu\end{vmatrix}}
  - \frac{\betat}{2}\psi_{0} \nonumber \\
  &- \frac{1}{2}\mathbf{y}_{d}^{T}\left[\betat\mathbf{I}_N - \betat^{2}\bm{\Psi}_{1}\left(\betat\bm{\Psi}_{2}+\Kuu\right)^{-1}\bm{\Psi}_{1}^{T}\right]\mathbf{y}_{d}, \label{eqn:f_d}
\end{align}
and the sufficient statistics for the covariance kernels are
\begin{equation}
\small
  \psi_{0} \!=\! \mathrm{Tr}\!\left(\mathbb{E}_q[\Kff]\right),\,
  \mathbf{\Psi}_{1} \!=\! \mathbb{E}_q[\Kfu],\,
  \mathbf{\Psi}_{2} \!=\! \mathbb{E}_q[\Kuf\Kfu]. \label{eqn:kernel_stats}
\end{equation}
The $\tilde{\Box}$ notation used here resolves the dependence on $\bZ$; these terms denote that expectations wrt $\bZ$ are being taken such that
\begin{equation}
\betat = \mathbb{E}_{q(\mathbf{z}_d)}\!\!\left[ \beta_t^{[\mathbf{z}_d = t]} \right].
\end{equation}
We use a similar notation for the covariance kernels; in addition, the subscripts refer to the locations used to evaluate the covariance functions with $\mathrm{f}$ denoting $\bX$ and $\mathrm{u}$ denoting $\Xu$. 
Thus, for example,
\begin{equation}
\Kfu = \mathbb{E}_{q(\mathbf{z}_d)}\!\!\left[ {k(\bX, \Xu ; \bm{\theta}_t)}^{[\mathbf{z}_d = t]} \right]. %
\end{equation}

\paragraph{DP Approximating Distributions} As specified previously, we use a stick breaking construction of the DP and introduce variational distributions over the assignment variables $\bZ$, the stick lengths $\bV$ and the clustering parameter $\alpha$ as a factorized distribution $q(\bZ,\bV,\alpha) \!=\! q(\bZ)q(\bV)q(\alpha)$ in a similar manner to~\cite{dp}. 

In order to deal with the infinite support of the DP, we artificially truncate the number of components to $T < \infty$. We note that this is not a particular limitation of our approach since in general the number of grouped functions will not exceed the number of observed dimensions $D$. 

In the truncated stick-breaking representation, it is assumed that the likelihood of the length of the stick drawn at $T$ is 1, therefore $q(v_T=1)=1$ and $\pi_t(\mathbf{V})=0$ for all $t > T$. This allows a finite approximating distribution to be used over the stick lengths; we use Beta distributions such that
\begin{equation}
q(\bV) \sim \prod_{t=1}^{T-1} \text{Beta}(v_t \vert a_t, b_t),  
\end{equation}
with $\bm{a}$ and $\bm{b}$ as variational parameters.

The truncation at $T$ also allows us to use a parameterized Multinomial for the approximate distribution over $\bZ$ as
\begin{equation}
q(\bZ) \sim  \prod_{d=1}^{D} \text{Mult}(\mathbf{z}_d \vert \bm{\phi}_d), \; \sum_{t=1}^{T} \phi_{d,t} = 1.
\end{equation}
For the cluster parameter we introduce a Gamma approximating distribution
$q(\alpha) = \text{Gamma}(\alpha \vert w_1, w_2)$.

\paragraph{DP Bound} These approximations lead to a tractable DP lower bound of 
\begin{multline}
\small  \mathcal{L}_{\mathcal{DP}} = \sum_{d=1}^{D}{\Big[\mathbb{E}_q[\log p(\mathbf{z}_d \vert \mathbf{V})]\Big]} + \mathbb{E}_q[\log p(\mathbf{V} \vert \alpha)] \\
\small  + \mathbb{E}_q[\log p(\alpha)] + \mathbb{H}[q(\mathbf{V})] + \mathbb{H}[q(\mathbf{Z})] + \mathbb{H}[q(\alpha)], \label{eqn:dp_bound}
\end{multline}
where all terms are defined over standard exponential family distributions.

\paragraph{Optimisation} We optimize the objective of~\eqref{eqn:objective} wrt the variational parameters $\{\variationalParams\}$ and the hyperparameters and noise precisions $\{\bm{\Theta}, \bm{\beta}\}$. We initialise the mean parameters $\bm{\mu}$ for $q(\bX)$ with the first $Q$ principal components of the observed data $\bY$ and set all $\bm{\Sigma}_q = \bm{I}_N$.  The pseudo input locations $\Xu$ are initialised to a random subset of $\bm{\mu}$. The stick length parameters $\bm{a}$ and $\bm{b}$ are drawn from a standard log-Normal. The allocation parameters $\bm{\Phi}$ are drawn from a standard normal pushed through the soft-max function. The hyperparameters and noise precisions are initialized with draws from their log-Normal priors. Finally, the shape and scale parameters for the Gamma distribution over $\alpha$ are initialized to their prior $w_1 = s_1 = 1$, $w_2 = s_2 = 1$.
In our experiments, we evaluate the variational lower bound and optimize it directly in TensorFlow using gradient descent with momentum.

\paragraph{Appendices} The appendices contain further information about the potential use of alternative priors on the latent space (\S~\ref{sec:latent_prior}), the stable calculation of the GP lower bound~\eqref{eqn:f_d} (\S~\ref{sec:stable_calculation}), the calculation of the kernel expectations of~\eqref{eqn:kernel_stats} (\S~\ref{sec:kernel_expectations}) and the terms in the DP lower bound of~\eqref{eqn:dp_bound} (\S~\ref{sec:details_dp_bound}).

\paragraph{Prediction and Missing Data} After training, prediction from the latent space follows straight forwardly from the Bayesian GP-LVM~\citep{bayesian_gplvm} where each dimension $d$ takes the kernel parameters from their respective posterior distribution. Since the model is fully generative, imputation of missing data is also a simple task. The model can be trained neglecting the observations for the missing data and then their value can be predicted from the posteriors conditioned on the observed data. 

If we would like to infer the location on the latent manifold for a new observation $\bY^{*}$, we can add additional variational parameters for $q(\bX^{*})$ and optimize the lower bound for the new joint model $(\bY,\bY^{*})$. The ratio of the lower bounds of the joint model to the original  can be used to infer the probability of the new data, as described by~\cite{bayesian_gplvm}.

\section{Experiments} \label{sec:experiments}
We now describe the experimental evaluation of the proposed DP-GP-LVM model. We test the model on three different data-sets with the aim of providing intuition to the benefit of our approach in comparison with previous models. As a first experiment we create a synthetic data-set where the groupings are known. 
The data are generated by specifying a GP and using samples from the model as observations. The twenty dimensional data were generated from three latent variables, where the first ten observed dimensions covary with latent dimensions one and two, and the second ten observed dimensions covary with latent dimensions one and three creating two distinct groups. 

\begin{figure}[tb]
  \centering
  \subfigure[ARD Weights]{
    \centering
    \includegraphics[width=0.45\linewidth]{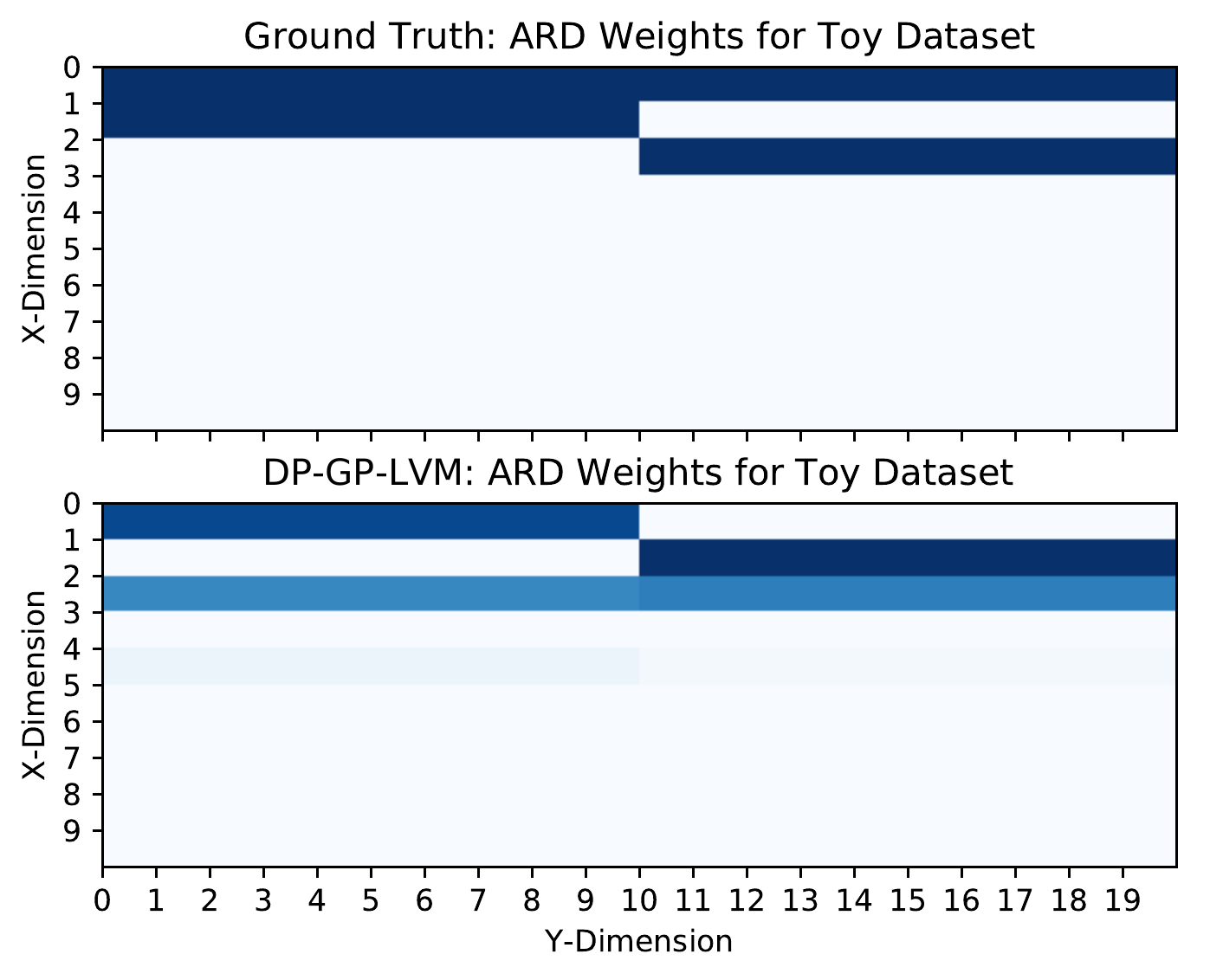}
    \label{fig:gp_dp_toy_ard}
  }\hfill
  \subfigure[DP Assigment]{
    \centering
    \includegraphics[width=0.45\linewidth]{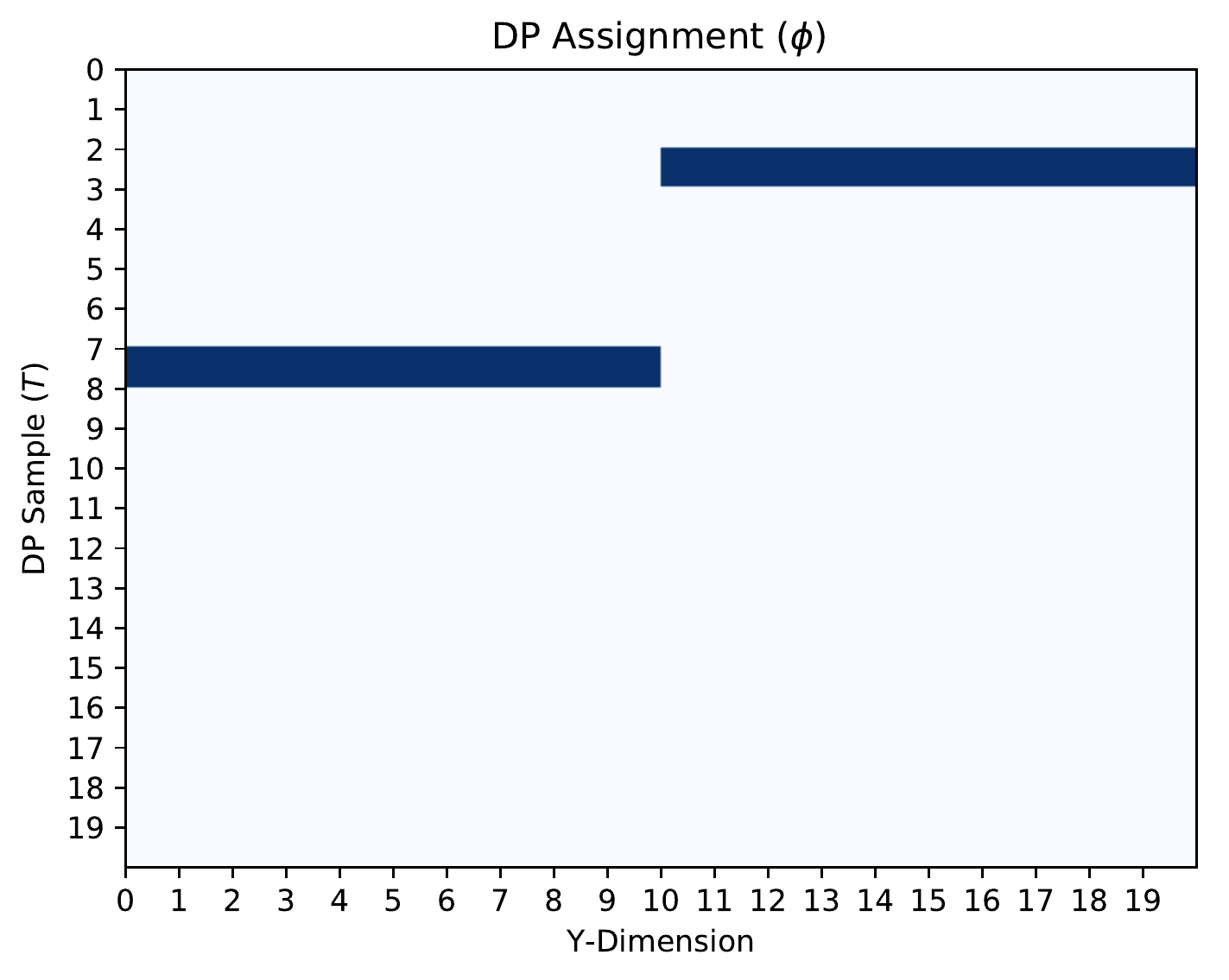}
    \label{fig:gp_dp_toy_phi}
  }
  \caption{DP-GP-LVM latent factorization of synthetic dataset. The twenty dimensional data have been generated from three latent variables, where the first ten observed dimensions covary with latent dimensions one and two, and the second ten observed dimensions covary with latent dimensions one and three creating two distinct groups. \subref{fig:gp_dp_toy_ard}~The top panel shows the ground truth grouping while the lower panel indicates the inferred grouping by DP-GP-LVM where a column indicates the latent dimensions in $\mathbf{X}$ allocated to represent the corresponding observed dimension $\mathbf{Y}$. \subref{fig:gp_dp_toy_phi}~The posterior DP allocations show that only two views are found and provides the correct allocations. Note the actual indices of the latent variables are not important due to the interchangeable characteristic of the DP prior.}
  \label{fig:toy_data_results}
\end{figure}

In Figure~\ref{fig:toy_data_results}, we can see that the DP-GP-LVM correctly recovers the latent structure underlying the creation of the data. Note the actual indices of the latent variables are not important due to the interchangeable characteristic of the DP prior.

\begin{figure*}[t!]
\setlength{\figwidth}{0.18\linewidth}
  \centering
    \subfigure[BGP-LVM]{%
        \centering
          \includegraphics[width=\figwidth]{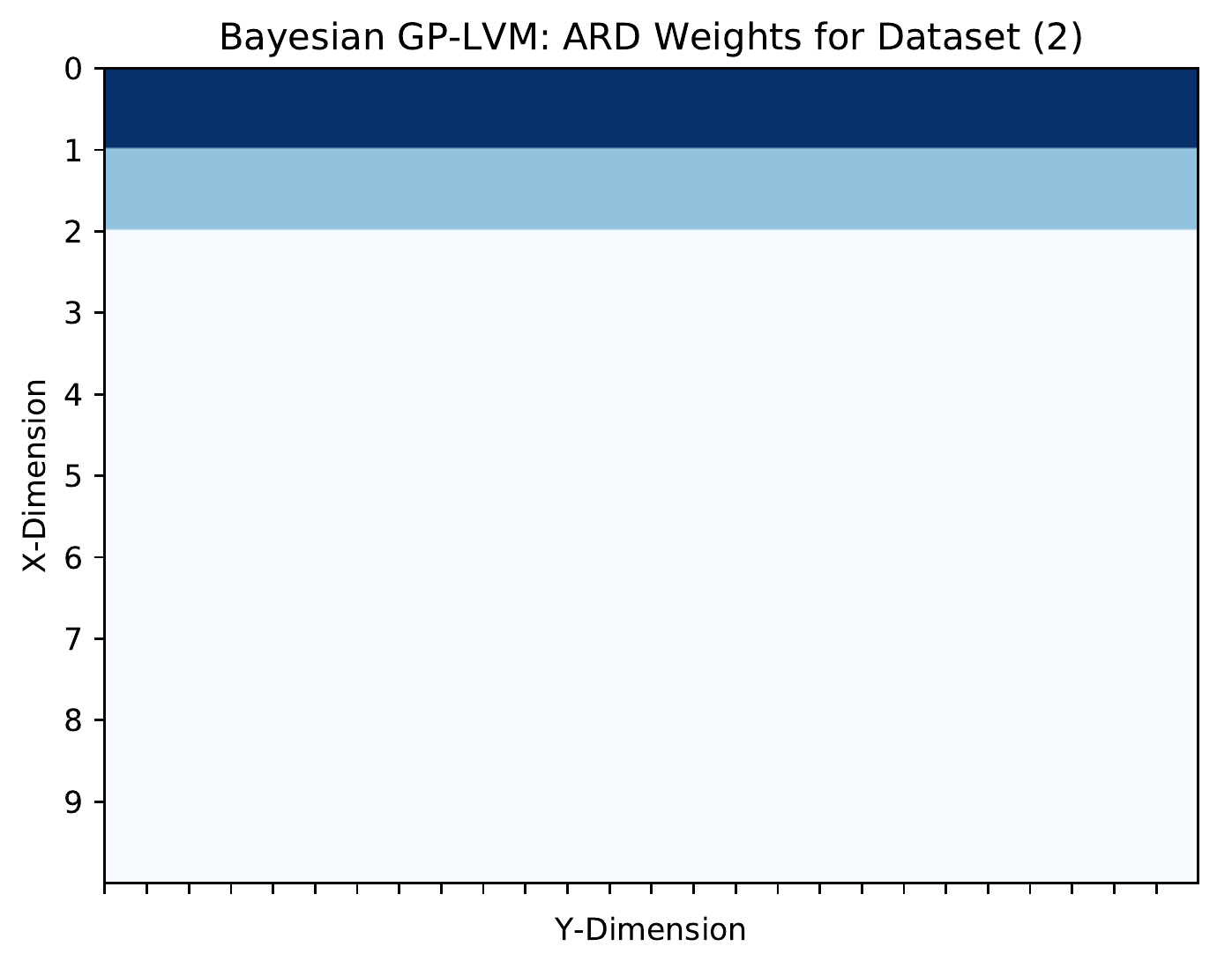}
          \label{fig:bayes_gplvm_ard_2}
    }\hfill
    \subfigure[MRD]{%
        \centering
          \includegraphics[width=\figwidth]{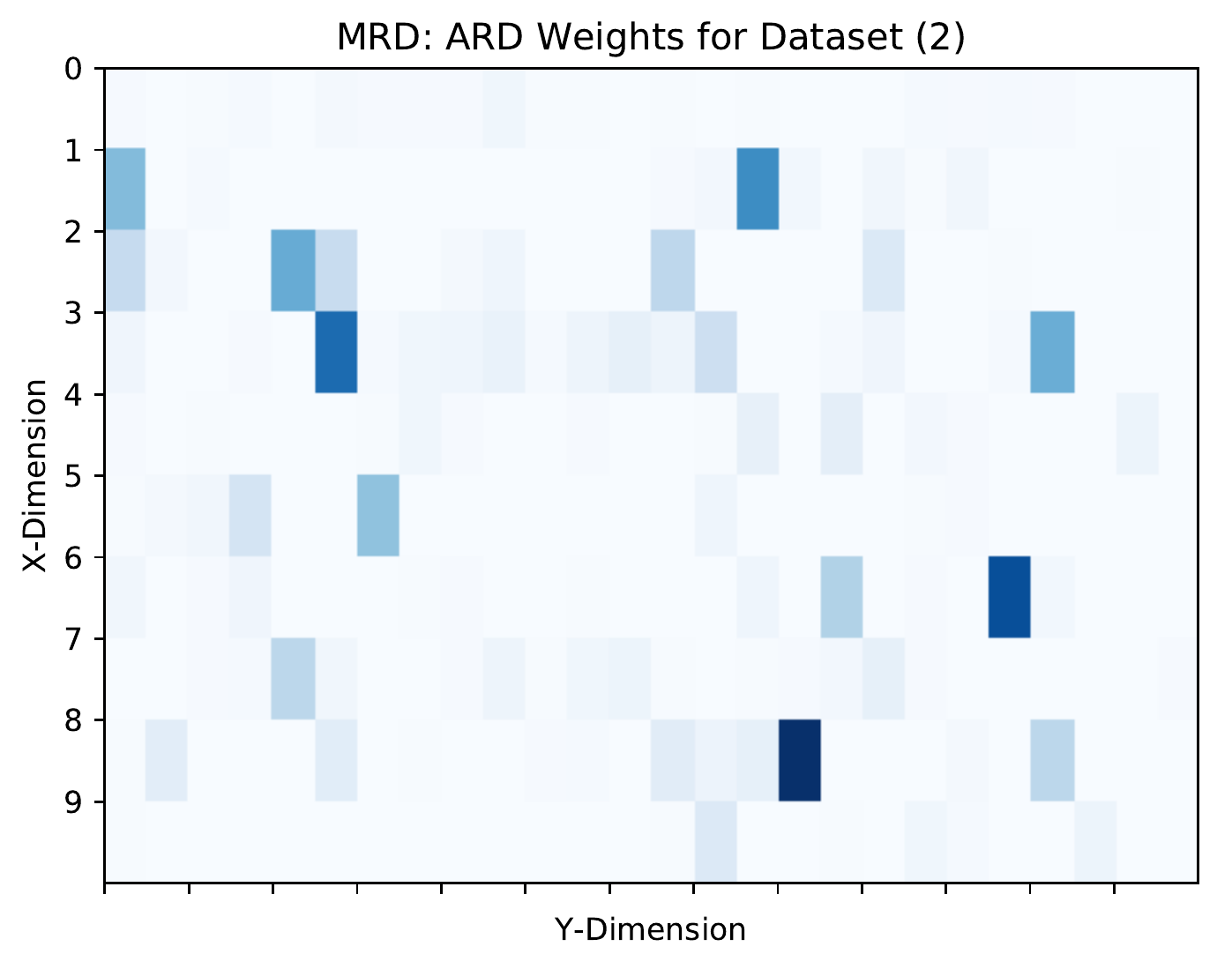}
          \label{fig:mrd_ard_2}
    }\hfill
    \subfigure[DP-GP-LVM]{%
        \centering
          \includegraphics[width=\figwidth]{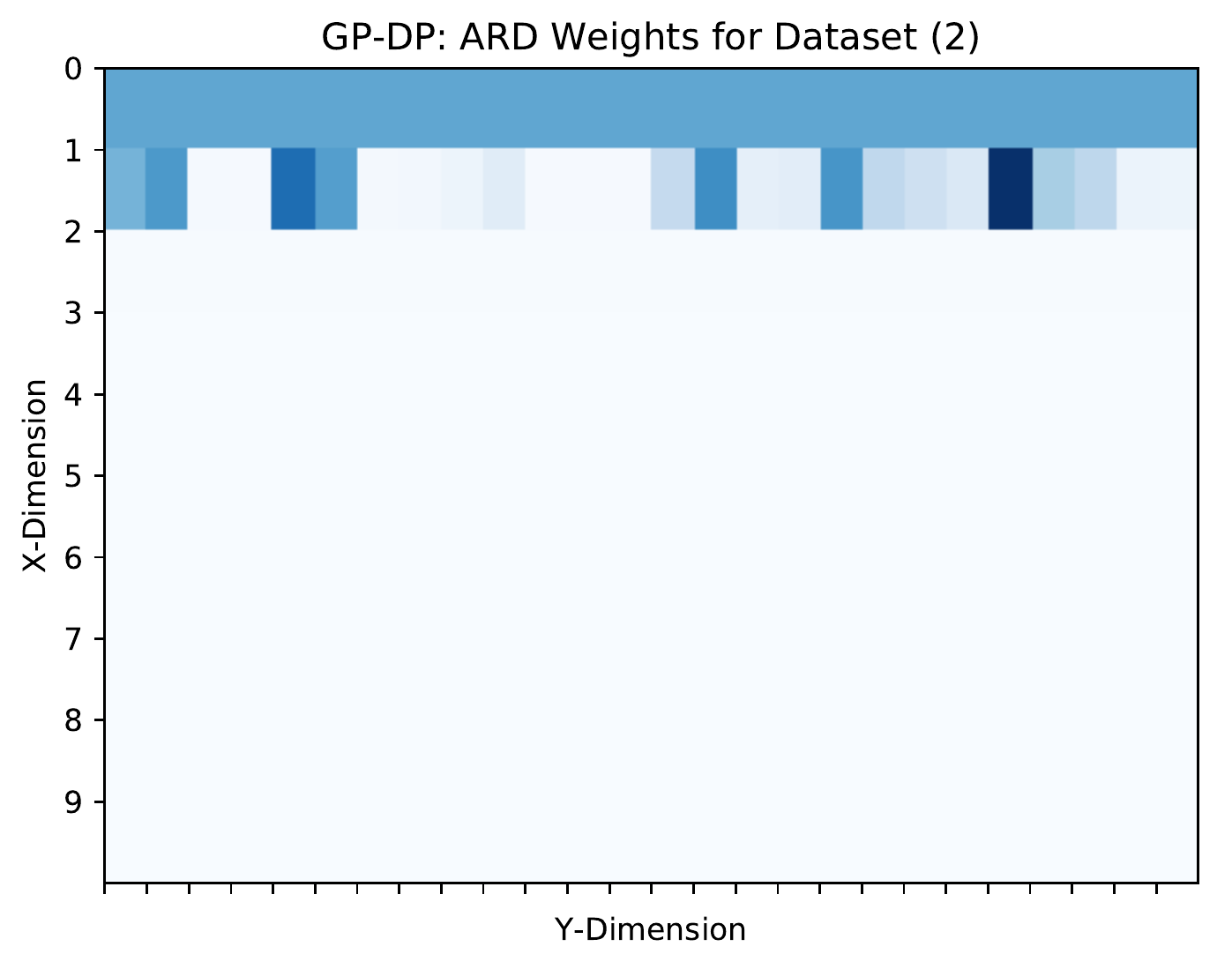}
          \label{fig:gp_dp_ard_2}
    }\hfill%
    \subfigure[BGP-LVM]{
        \centering
          \includegraphics[width=\figwidth]{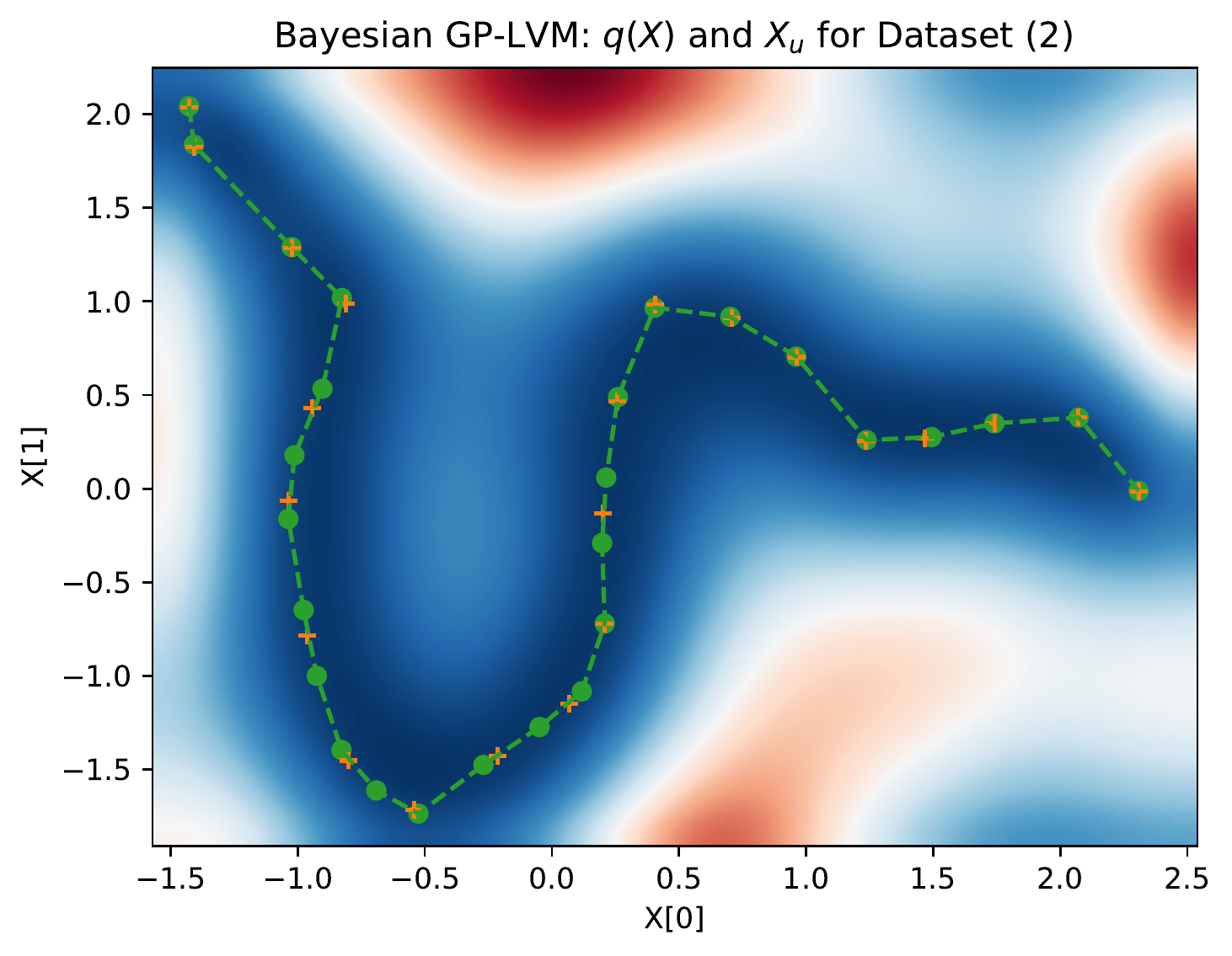}
          \label{fig:bayes_gplvm_latent01_2}
    }\hfill
    \subfigure[DP-GP-LVM]{
        \centering
          \includegraphics[width=\figwidth]{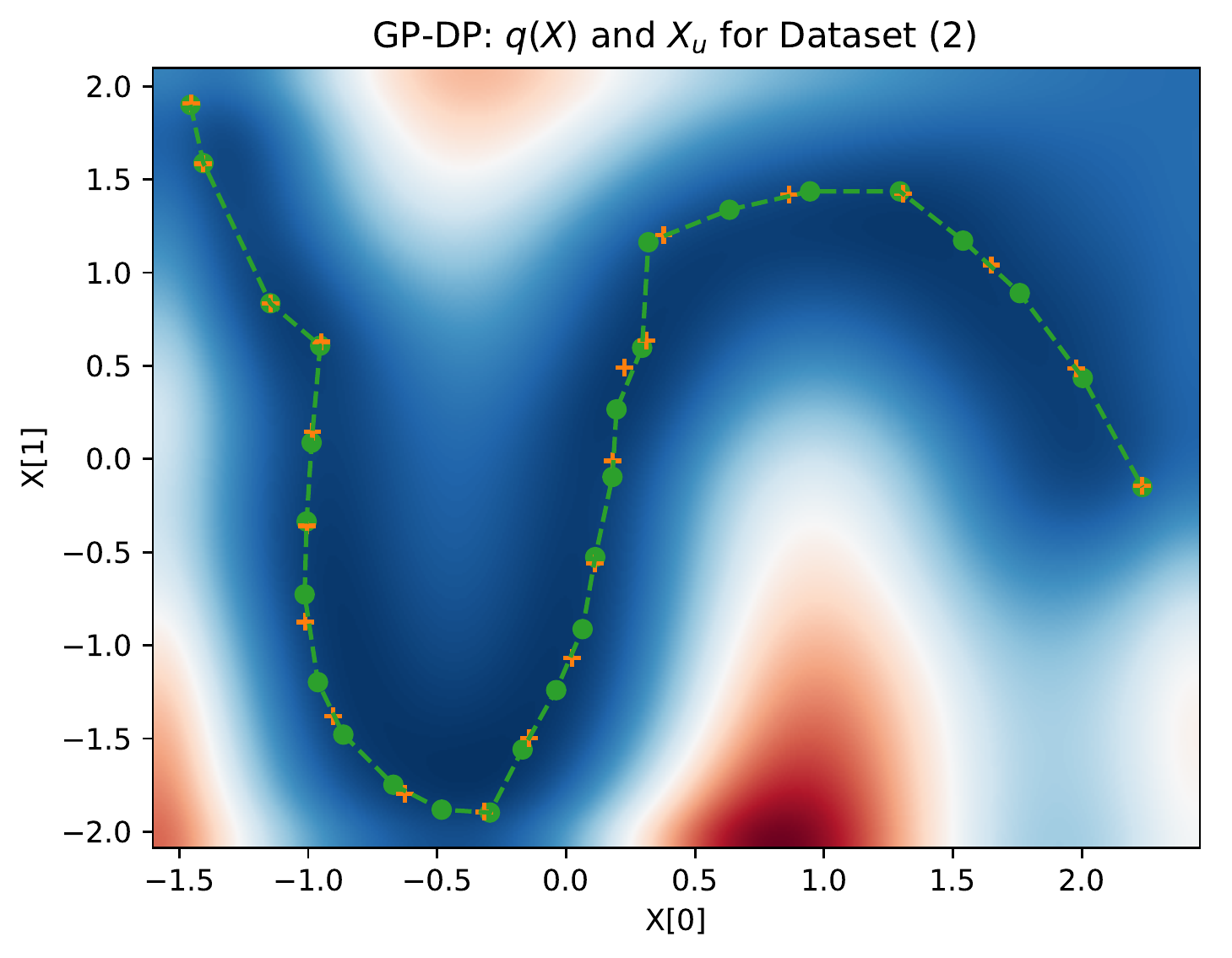}
          \label{fig:gp_dp_latent01_2}
    }~~~~\\[-10pt]%
    \caption{\small\it Results for PoseTrack dataset with two individuals.
      \subref{fig:bayes_gplvm_ard_2}-\subref{fig:gp_dp_ard_2} show the ARD weights returned for the three models. BGP-LVM result shares only two dimensions, failing to capture the full diversity of the data. MRD captures these changes but fails to find the appropriate correlations, creating unnecessary independence. DP-GP-LVM capture both the shared correlation and the subtle independences. \subref{fig:bayes_gplvm_latent01_2} and~\subref{fig:gp_dp_latent01_2} show the manifold for the first two shared latent dimensions for BGP-LVM and DP-GP-LVM indicating that they both capture the smooth structure.}
    \label{fig:pose_track_results_2}
\end{figure*}

\begin{figure*}[t!]
\setlength{\figwidth}{0.23\linewidth}%
  \centering
    \subfigure[BGP-LVM]{
        \centering
          \includegraphics[width=\figwidth]{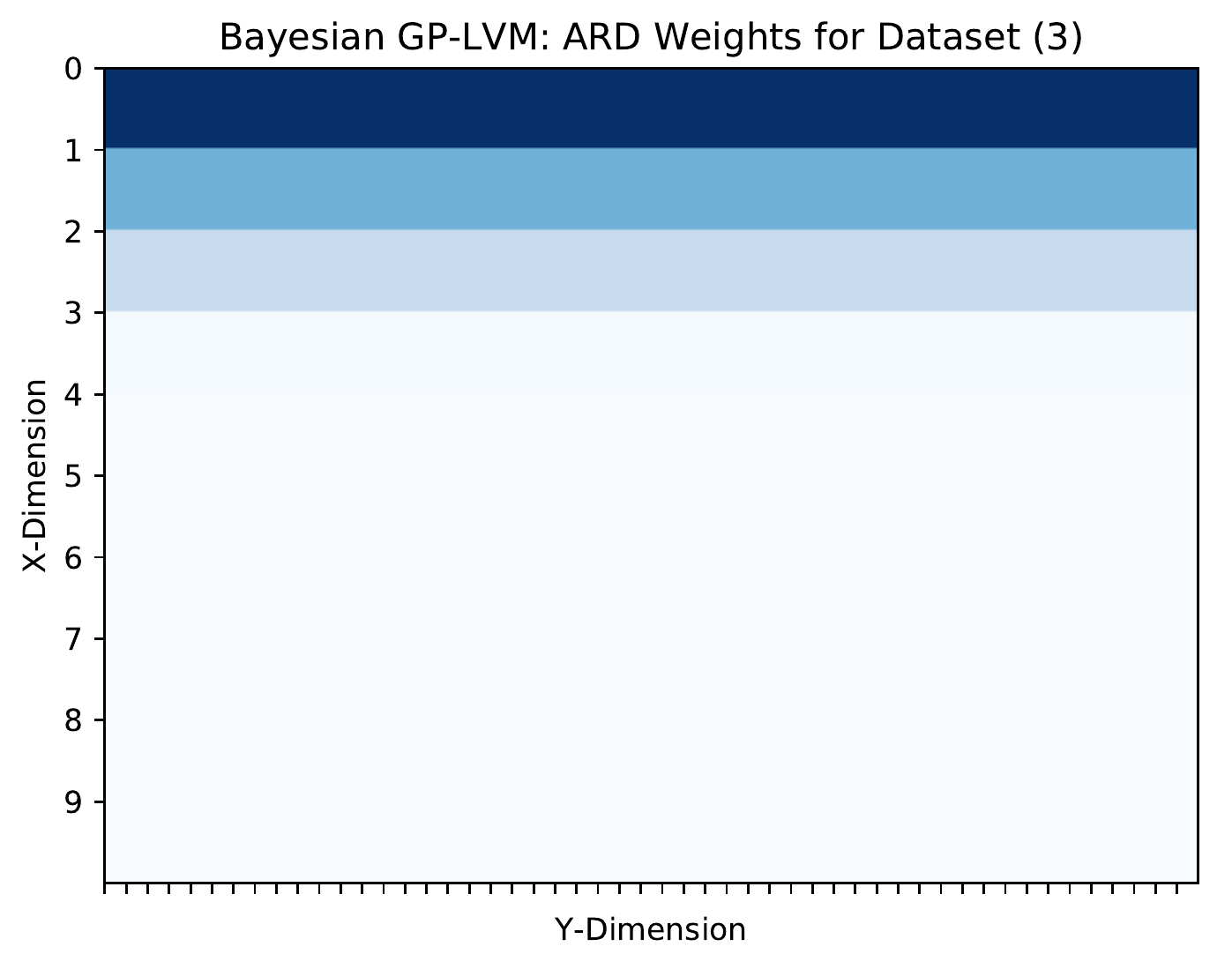}
          \label{fig:bayes_gplvm_ard_3}
    }\hfill
    \subfigure[MRD]{
        \centering
          \includegraphics[width=\figwidth]{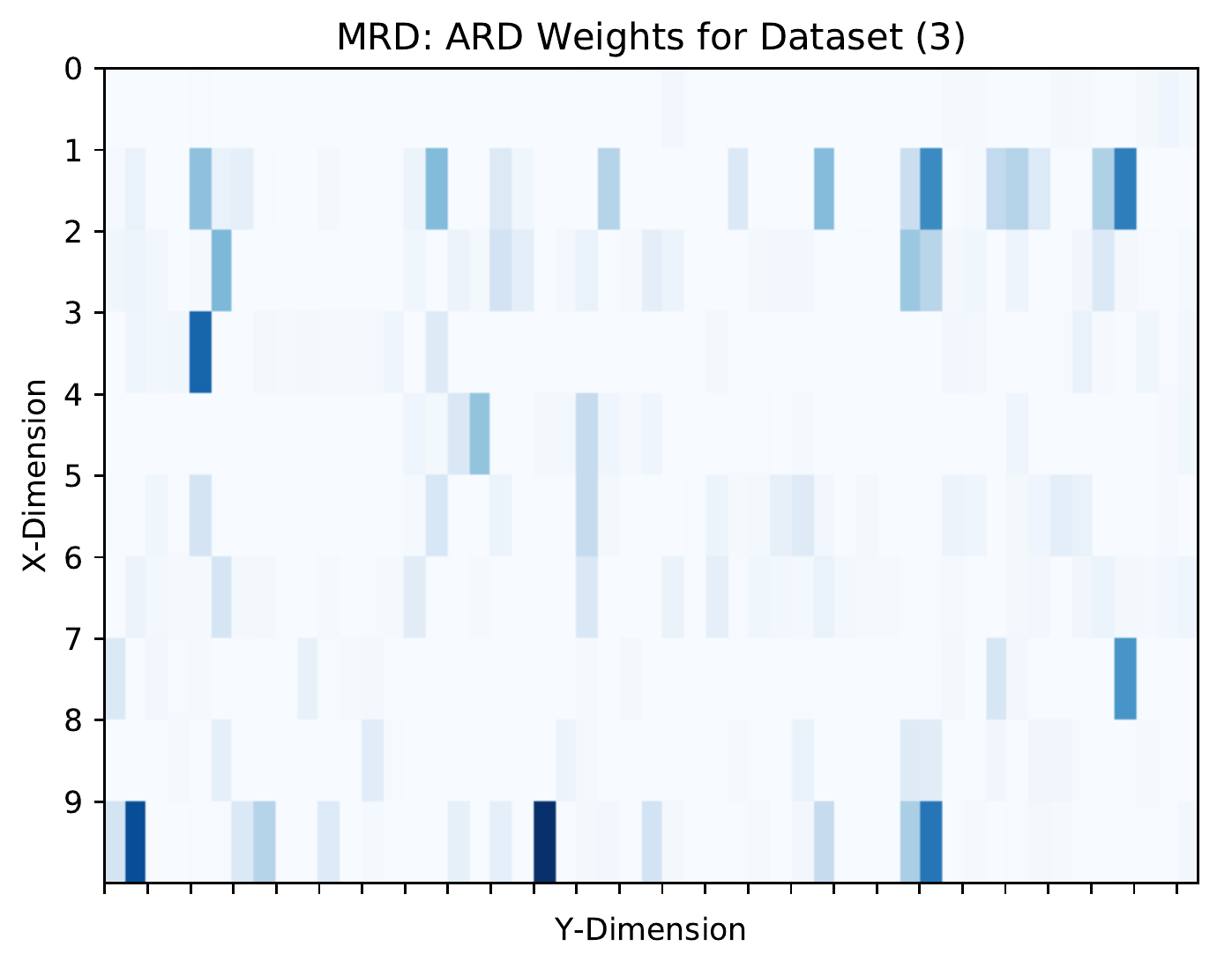}
          \label{fig:mrd_ard_3}
    }\hfill
    \subfigure[DP-GP-LVM]{
        \centering
          \includegraphics[width=\figwidth]{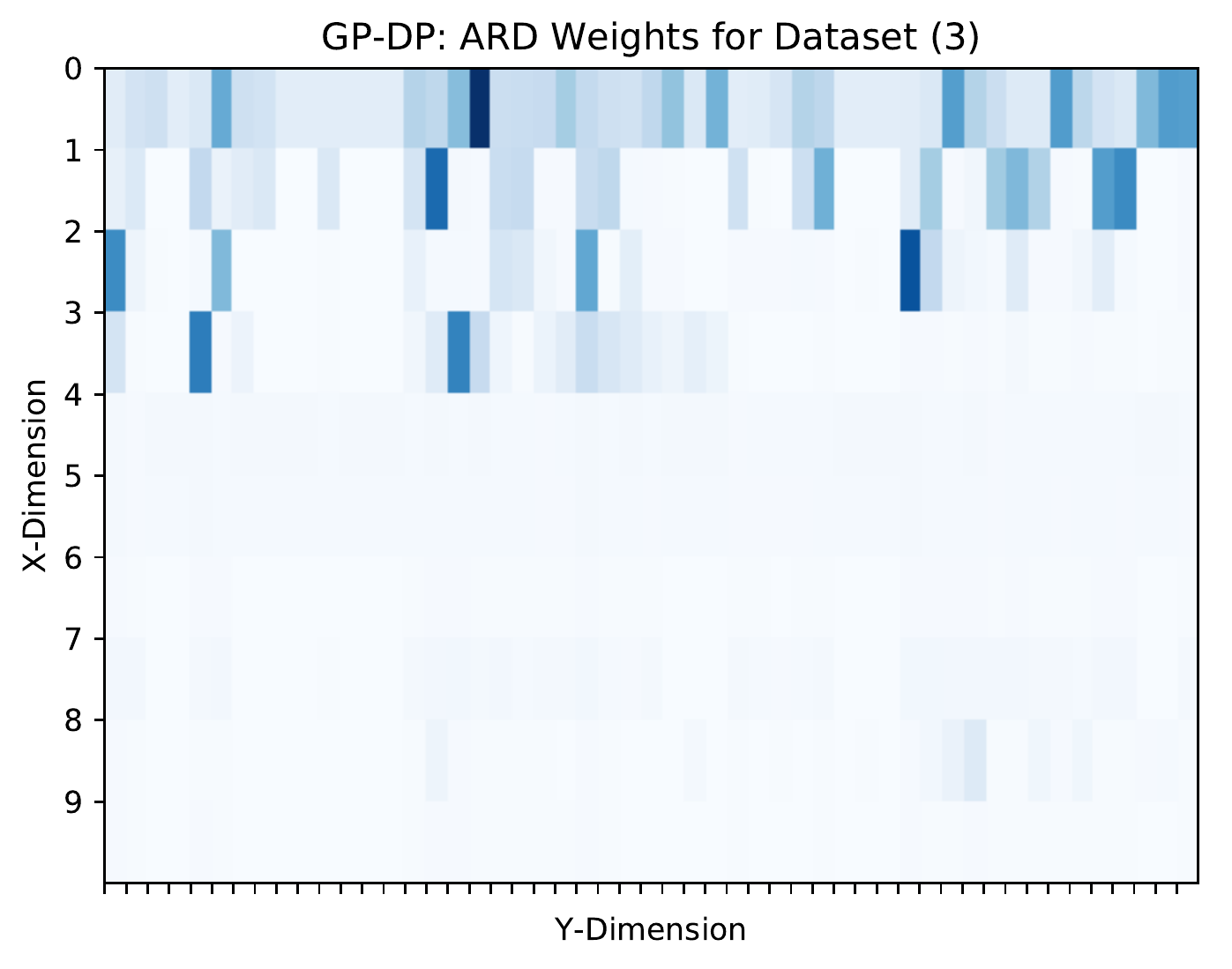}
          \label{fig:gp_dp_ard_3}
    }\hfill
    \subfigure[BGP-LVM]{
        \centering
          \includegraphics[width=\figwidth]{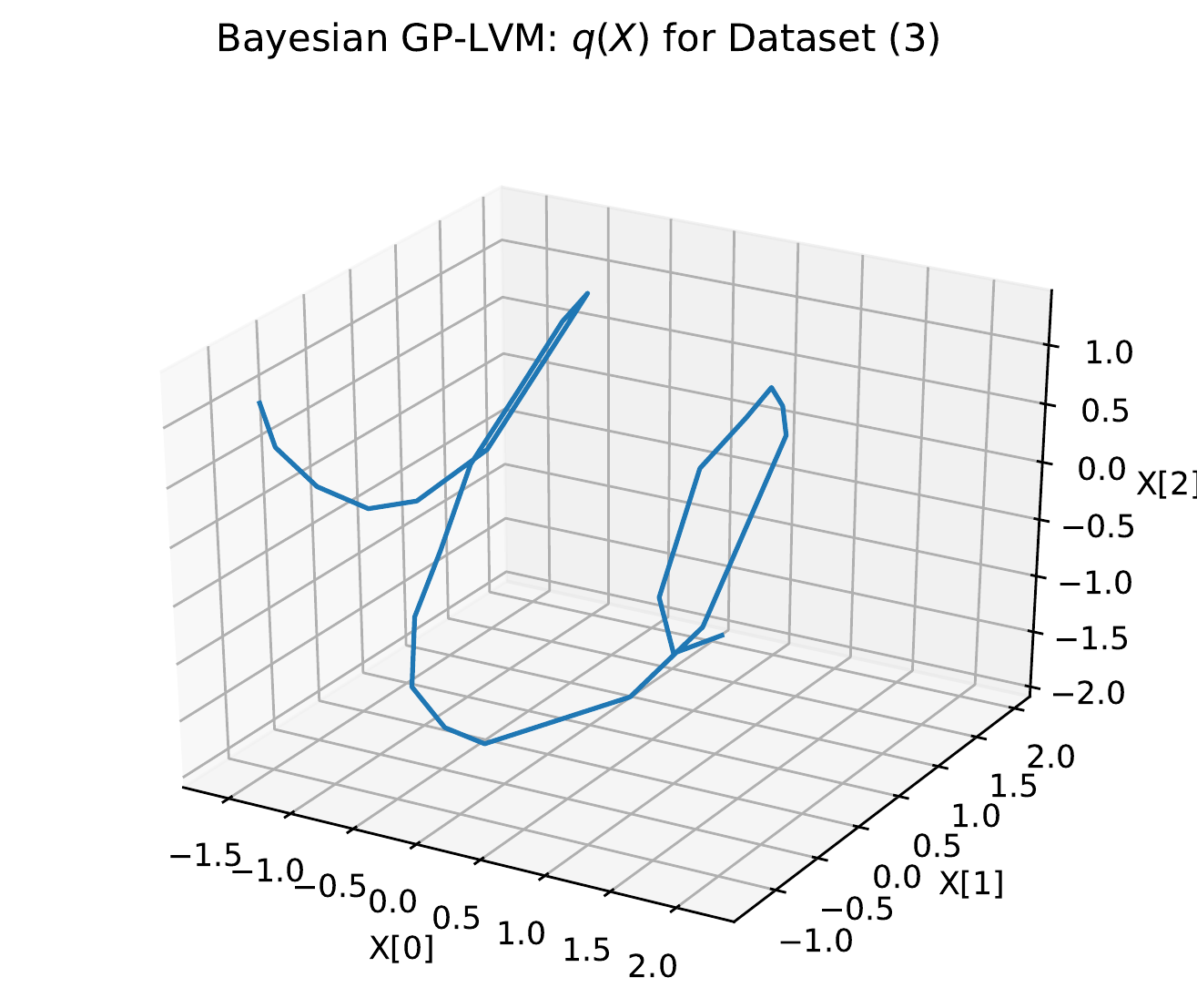}
          \label{fig:bayes_gplvm_latent3d_012_3}
    }\\[-2pt]%
    \setlength{\figheight}{0.23\linewidth}%
    \subfigure[DP-GP-LVM $X_1$-$X_3$]{
        \centering
          \includegraphics[height=\figheight]{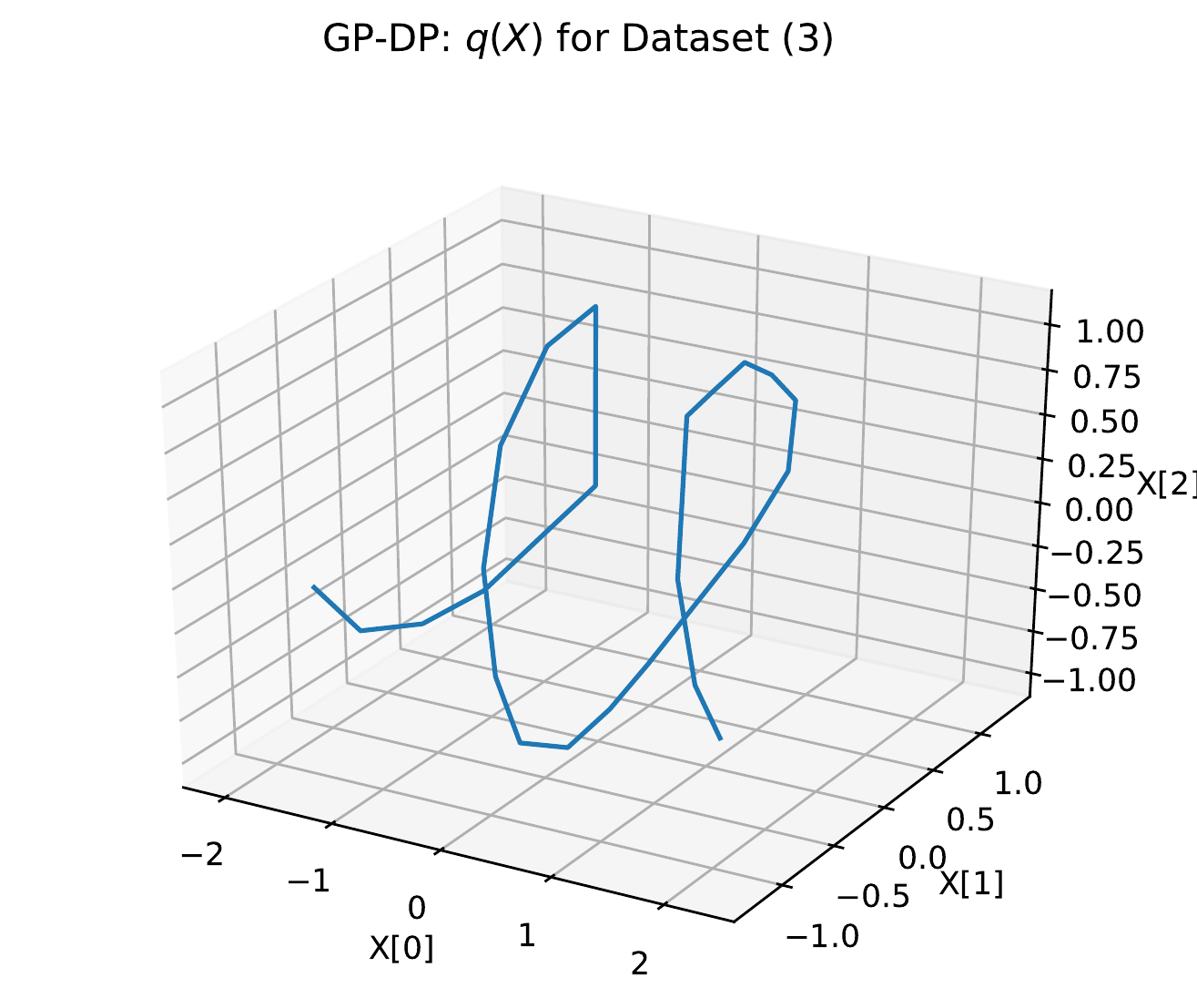}
          \label{fig:gp_dp_latent3d_012_3}
    }\hfill
    \subfigure[DP-GP-LVM $X_2$-$X_4$]{
        \centering
          \includegraphics[height=\figheight]{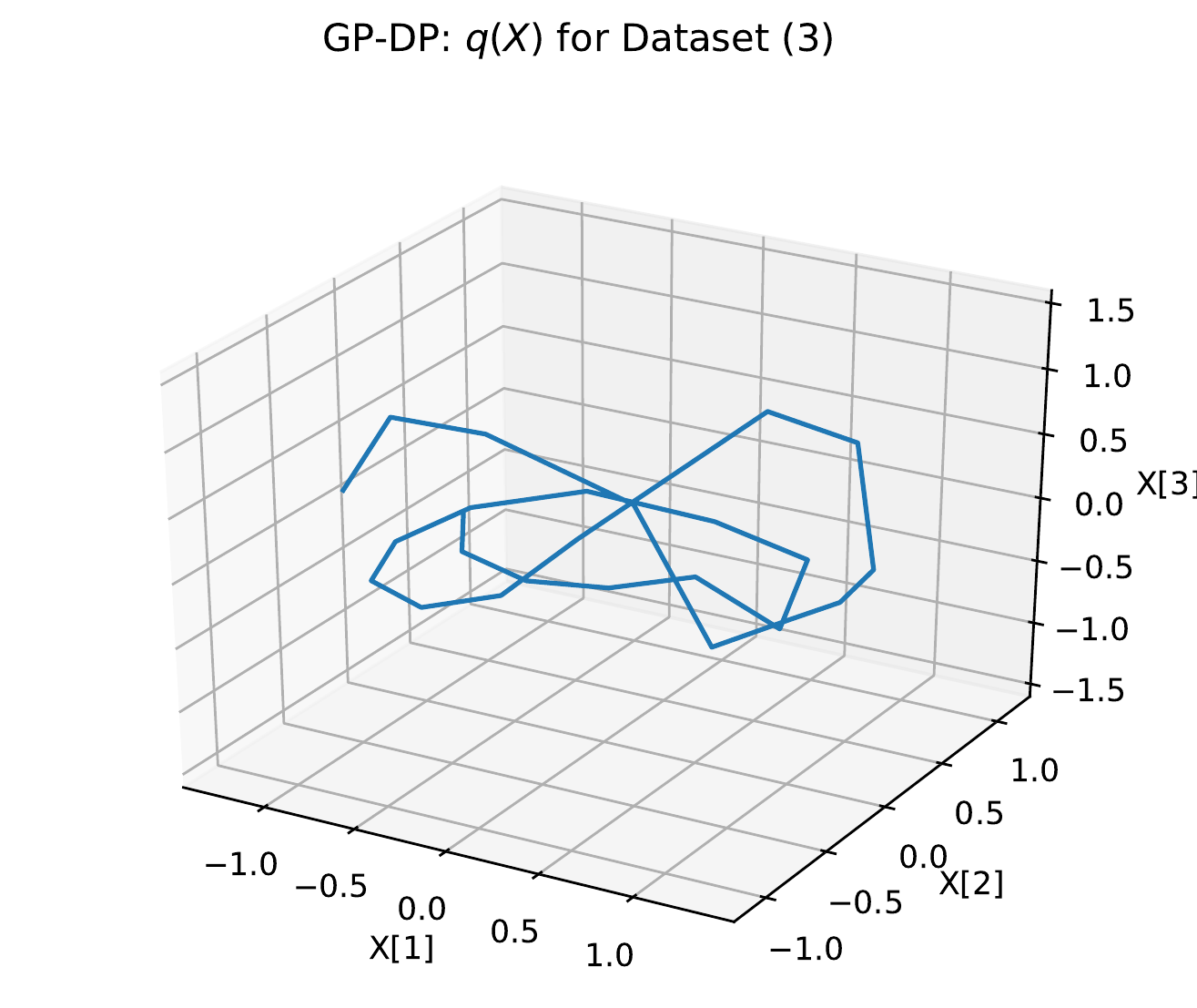}
          \label{fig:gp_dp_latent3d_123_3}
    }\hfill    \setlength{\figwidth}{0.45\linewidth}%
    \subfigure[t][DP-GP-LVM Inferred Groups]{
        \centering
          \includegraphics[height=\figheight]{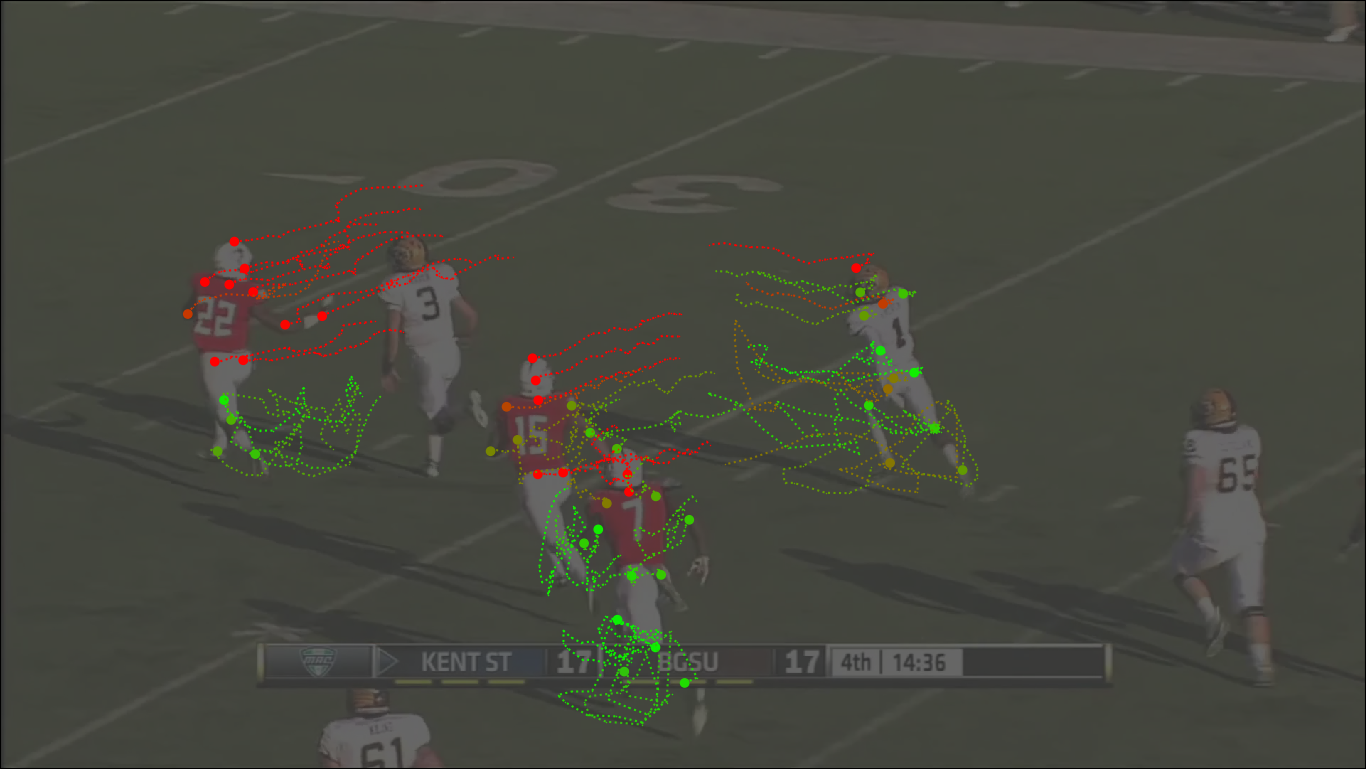}
          \label{fig:posetrack_image}
    }\\[-10pt]%
    \caption{\small\it Results for PoseTrack dataset with four individuals. \subref{fig:bayes_gplvm_ard_3}-\subref{fig:gp_dp_ard_3} the ARD weights found for the three models. BGP-LVM increases to three latent dimensions but without the independence structure captured by DP-GP-LVM.
    \subref{fig:bayes_gplvm_latent3d_012_3} and~\subref{fig:gp_dp_latent3d_012_3} show the smooth, interpretable manifolds from DP-GP-LVM with subtle structure that extends into the fourth latent dimension~\subref{fig:gp_dp_latent3d_123_3}. \subref{fig:posetrack_image}~Here we show the learned groups (posterior over $\bZ$) as the color of the points overlaid on the first frame of the image. The path traces out the future motion. We observe the separation of the clusters into translation and periodic motions.   }
    \label{fig:pose_track_results_3}
\end{figure*}

\begin{figure*}[tb]
\setlength{\figwidth}{0.28\linewidth}
  \centering
    \subfigure[BGP-LVM]{
        \centering
          \includegraphics[width=\figwidth]{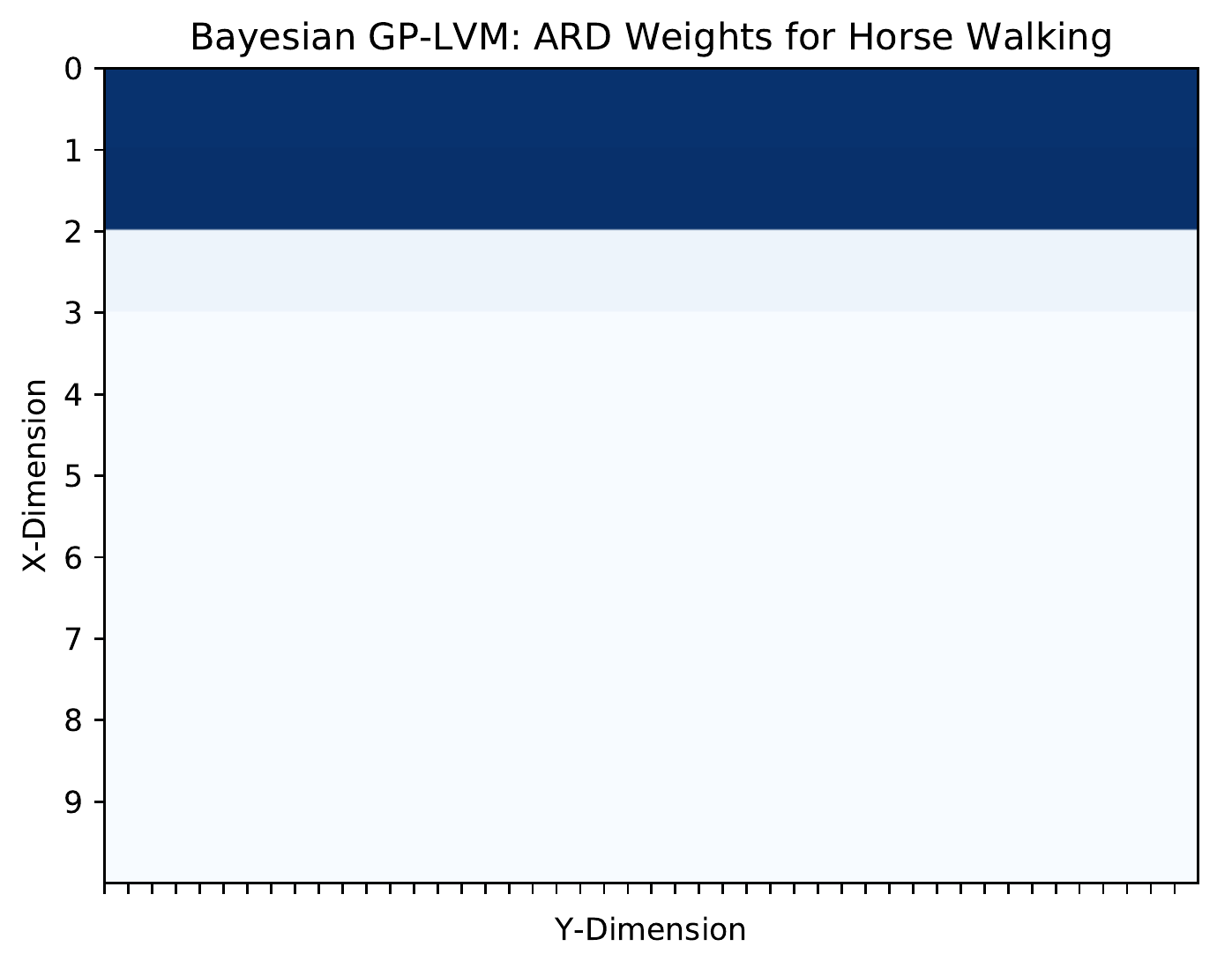}
          \label{fig:bayes_gplvm_ard_horse_walking}
    }\hfill
    \subfigure[MRD]{
        \centering
          \includegraphics[width=\figwidth]{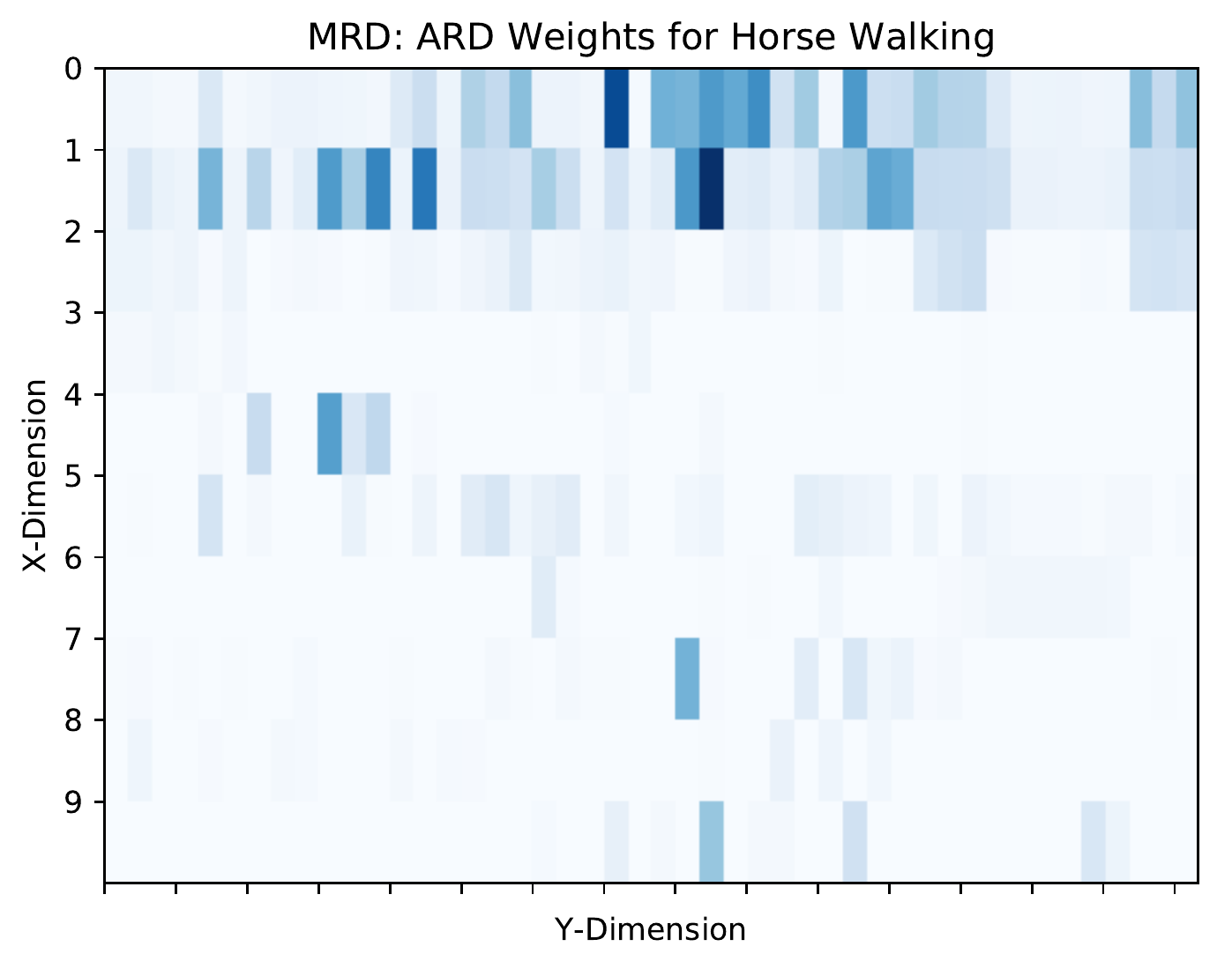}
          \label{fig:mrd_ard_horse_walking}
    }\hfill
    \subfigure[DP-GP-LVM]{
        \centering
          \includegraphics[width=\figwidth]{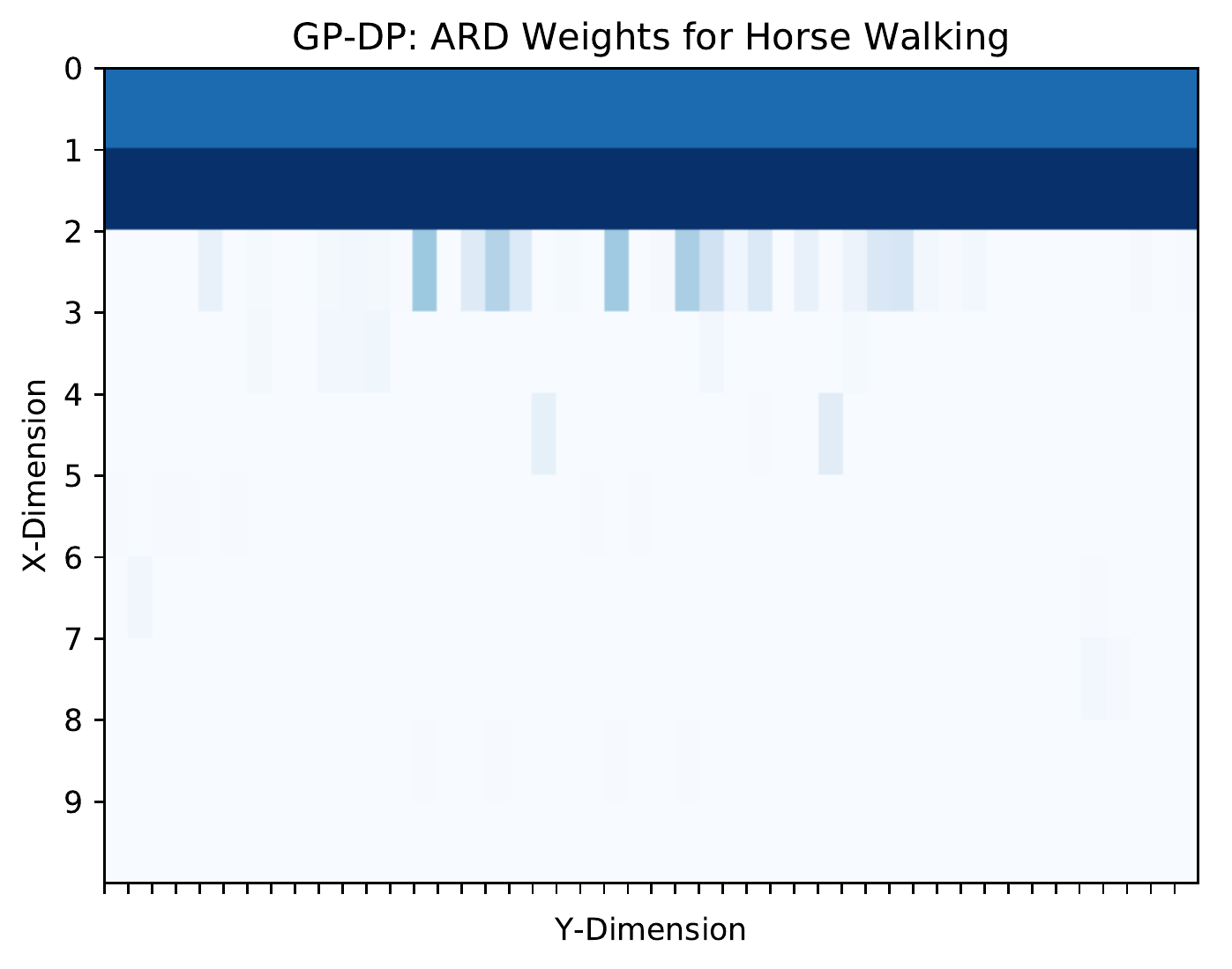}
          \label{fig:gp_dp_ard_horse_walking}
    }\\[-5pt]%
\setlength{\figwidth}{0.45\linewidth}%
    \subfigure[Effect of $X_3$]{
        \centering
          \includegraphics[width=\figwidth]{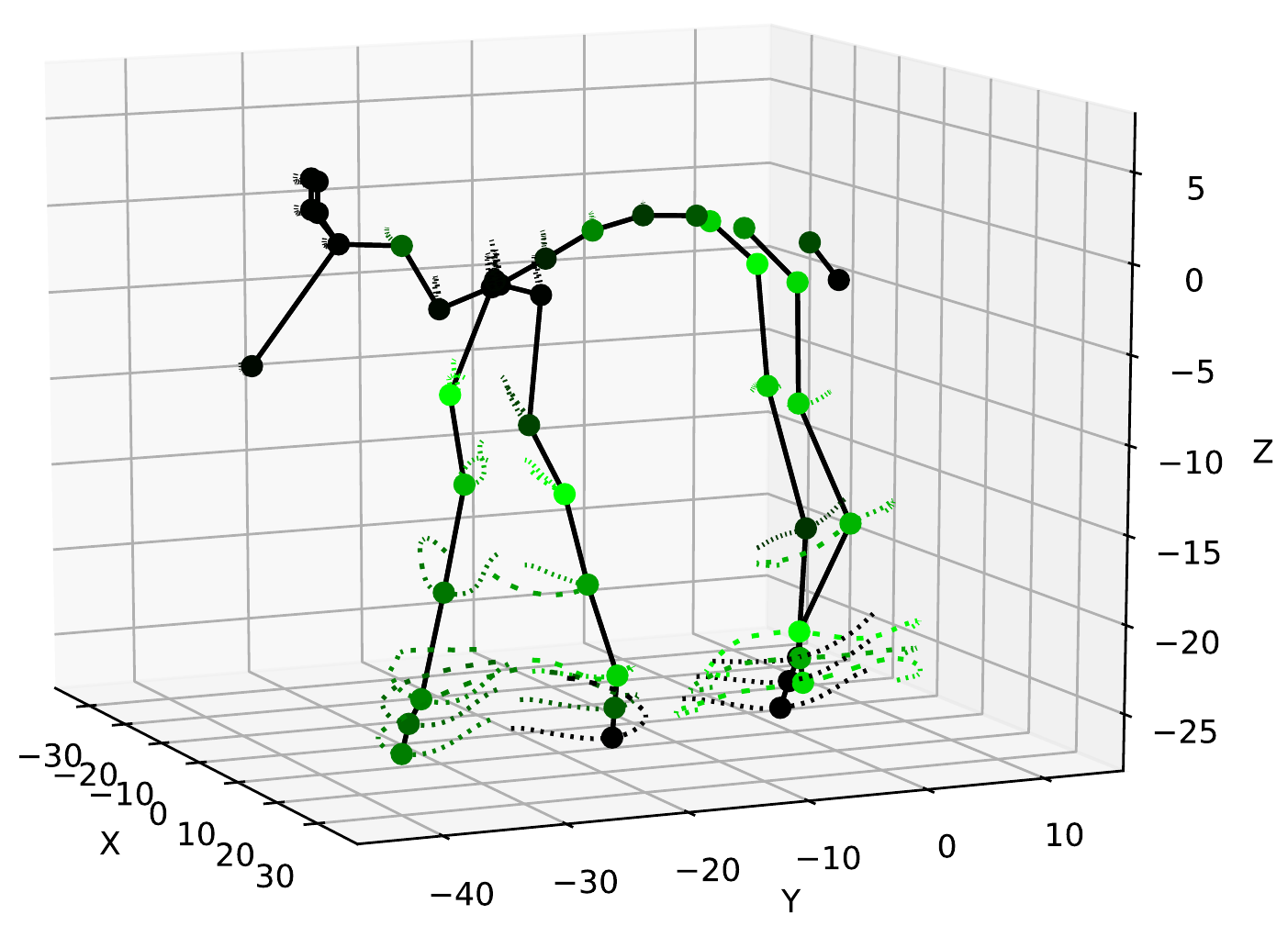}
          \label{fig:horse_walking}
    }\hfill
    \subfigure[DP-GP-LVM $X_1$-$X_3$]{
        \centering
          \includegraphics[width=\figwidth]{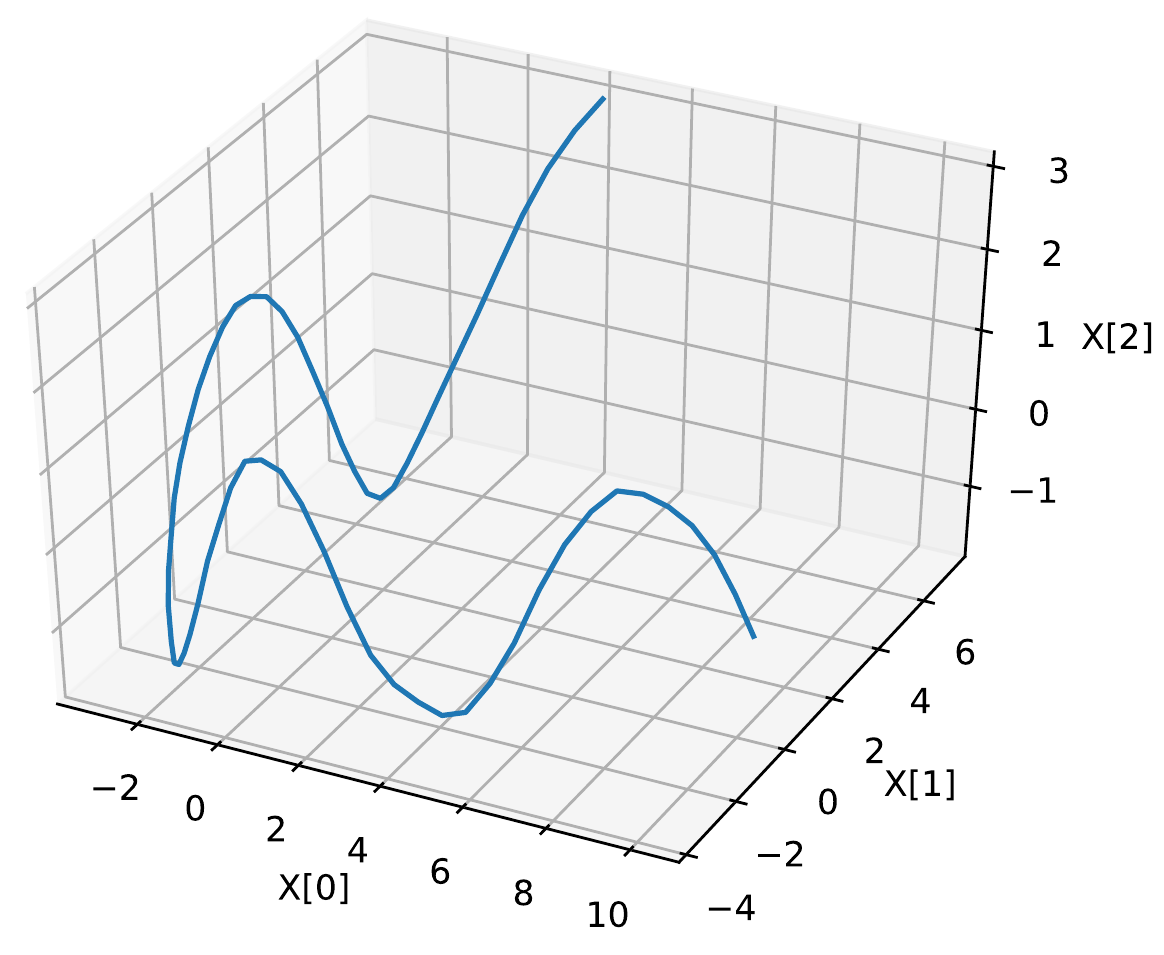}
          \label{fig:gp_dp_latent3d_horse_walking}
    } \\[-10pt]%
    \caption{\small\it Horse Walking Dataset. \subref{fig:bayes_gplvm_ard_horse_walking}-\subref{fig:gp_dp_ard_horse_walking}  ARD weights for the three models. BGP-LVM can model the data well with two latent dimensions %
    while MRD is unable to learn a reasonable factorization of the latent space. 
    DP-GP-LVM uses three dimensions with the first two shared and the third encoding the oscillating motion at certain joints when the horse is walking. \subref{fig:horse_walking} The position of the horse at a frame in the middle of the sequence. Each joint has a line showing the position of the joint in past and future frames with the color modulated by the third ARD weight. %
   Joints in the legs are a function of $X_3$, while the spine and head are not, due to the legs oscillating as the horse is walking. \subref{fig:gp_dp_latent3d_horse_walking}~The path in the latent space for DP-GP-LVM is periodic.}
    \label{fig:horse_walking_results}
\end{figure*}

\paragraph{Motion Capture Datasets} A motivation to our work is to avoid the requirement of defining groupings a priori. While we believe that the goal of learning this directly from data is justified, it creates challenges in terms of qualitative evaluation. To this end, we focus on learning representations of motion capture data as it provides interpretable correlation structures. 

PoseTrack~\citep{PoseTrack} consists of spatial image locations corresponding to an underlying human three dimensional motion. We create three separate data sets corresponding to the motion of two and four individuals as this will allow us to evaluate groupings both within and between individuals. We compare the DP-GP-LVM with a model where the observed data is considered to be a single group (BGP-LVM) and a model where each dimension is considered its own group (MRD). Importantly, our model contains both these two cases but marginalises over them and all other combinations. 

Figure~\ref{fig:pose_track_results_2} shows the results for two individuals. The MRD model where each dimension is a group fails to capture the correlation structure and creates a large number of groups. The single group and the DP-GP-LVM both use a two dimensional latent representation, however, where the former is forced to use both latent dimensions across all dimensions the latter is able to only use the a second dimension for a subset of the observed variates. When looking at the latent representation that has been recovered the DP-GP-LVM is capable of learning a smoother representation as the additional latent dimension is not ``polluting'' the latent space as with the single group model. 

Increasing the number of individuals to four (see Figure~\ref{fig:pose_track_results_3}) the characteristics of the solutions changes where the DP-GP-LVM now includes four dimensions while the single group model only uses three. The subtle variations which the DP-GP-LVM captures with its fourth dimension will be explained away as noise in the single view model as it is not present in a majority of the dimensions.

In Figure~\ref{fig:posetrack_image} we show the mapping of the groupings of the joints superimposed onto the image. As can be seen, the model has grouped joints on the upper-body, which have mainly translational variation, separately from the lower-body which, in addition, have significant oscillation due to the leg movement. There is also a third group which is only present in one of the individuals corresponding to a difference in translational movement.

As a third and final data-set, we apply the model to a three dimensional motion sequence of a horse~\citep{horseMocap}. The skeleton consists of $46$ joints leading to $138$ observed dimensions. To show the models ability to learn from small amounts of data we use only $63$ time-steps in a sequence. In Figure~\ref{fig:horse_walking_results} the results are shown. Again, we see that the BGP-LVM over simplifies the structure and MRD over complicates it. Inspection of the inferred structure, in Figure~\ref{fig:horse_walking}, confirms the main differences in the grouping arise from the periodic motion of the limbs compared to the head and torso.

\paragraph{Missing Data Experiment} We also performed a quantitative comparison for imputing missing data for the PoseTrack with four individuals. Table~\ref{tab:missing_data_results} shows the MSE between the predictions and ground truth for the missing data under each model. The data was generated by randomly removing a quarter of the dimensions and data points from the dataset. 
\begin{table}[t!]
  \centering
  \scalebox{0.6}{%
  \centering
    \begin{tabular}{ccccccccc}
      \toprule
      & \multicolumn{2}{c}{\textbf{Missing (\%)}} & \multicolumn{2}{c}{\textbf{MSE}} & \multicolumn{2}{c}{\textbf{Opt. Time (s)}} \\ \cmidrule(r{1pt}){2-3} \cmidrule(r{1pt}){4-5} \cmidrule(r{1pt}){6-7} \cmidrule(l{1pt}){8-9}
      & N & D & 2 Person & 4 Person & 2 Person & 4 Person  \\
      \midrule
      BGP-LVM & \multirow{3}{*}{$25$} & \multirow{3}{*}{$10$} &
        $0.06 \pm 0.01$ & %
        $0.11 \pm 0.03$ & %
        $\phantom{0}10.1 \pm 0.3$ & %
        $\phantom{0}11.0 \pm \phantom{0}0.6$ \\ %
      MRD &&&
        $0.07 \pm 0.01$ & %
        $0.11 \pm 0.03$ & %
        $172.8 \pm 2.5$ & %
        $370.7 \pm 18.5$ \\ %
      DP-GP-LVM &&&
        $0.06 \pm 0.01$ & %
        $0.10 \pm 0.02$ & %
        $\phantom{0}80.5 \pm 1.1$ & %
        $126.0 \pm \phantom{0}3.6$ \\ %
      \midrule
      BGP-LVM & \multirow{3}{*}{$25$} & \multirow{3}{*}{$25$} &
        $0.07 \pm 0.01$ & %
        $0.10 \pm 0.02$ & %
        $\phantom{0}10.4 \pm 0.3$ & %
        $\phantom{0}11.7 \pm \phantom{0}0.6$ \\ %
      MRD &&&
        $0.10 \pm 0.02$ & %
        $0.11 \pm 0.02$ & %
        $175.0 \pm 5.6$ & %
        $363.2 \pm 22.5$ \\ %
      DP-GP-LVM &&&
        $0.07 \pm 0.01$ & %
        $0.10 \pm 0.02$ & %
        $\phantom{0}77.3 \pm 0.9$ & %
        $120.1 \pm \phantom{0}4.3$ \\ %
      \midrule
      BGP-LVM & \multirow{3}{*}{$25$} & \multirow{3}{*}{$50$} &
        $0.07 \pm 0.01$ & %
        $0.09 \pm 0.01$ & %
        $\phantom{0}10.4 \pm 0.3$ & %
        $\phantom{0}11.9 \pm \phantom{0}0.7$ \\ %
      MRD &&&
        $0.11 \pm 0.01$ & %
        $0.13 \pm 0.02$ & %
        $171.9 \pm 2.9$ & %
        $360.9 \pm 20.7$ \\ %
      DP-GP-LVM &&&
        $0.07 \pm 0.01$ & %
        $0.09 \pm 0.01$ & %
        $\phantom{0}79.8 \pm 0.3$ & %
        $123.4 \pm \phantom{0}3.3$ \\ %
      \midrule
      BGP-LVM & \multirow{3}{*}{$50$} & \multirow{3}{*}{$10$} &
        $0.20 \pm 0.09$ & %
        $0.24 \pm 0.06$ & %
        $\phantom{0}11.1 \pm 0.2$ & %
        $\phantom{0}12.6 \pm \phantom{1}0.5$ \\ %
      MRD &&&
        $0.23 \pm 0.09$ & %
        $0.27 \pm 0.05$ & %
        $180.9 \pm 2.1$ & %
        $363.1 \pm 14.7$ \\ %
      DP-GP-LVM &&&
        $0.19 \pm 0.08$ & %
        $0.24 \pm 0.06$ & %
        $\phantom{0}82.9 \pm 0.3$ & %
        $123.4 \pm \phantom{1}3.3$ \\ %
      \midrule
      BGP-LVM & \multirow{3}{*}{$50$} & \multirow{3}{*}{$25$} &
        $0.21 \pm 0.07$ & %
        $0.22 \pm 0.03$ & %
        $\phantom{0}11.1 \pm 0.3$ & %
        $\phantom{0}12.5 \pm \phantom{1}0.5$ \\ %
      MRD &&&
        $0.22 \pm 0.07$ & %
        $0.25 \pm 0.04$ & %
        $176.8 \pm 4.9$ & %
        $354.1 \pm 12.3$ \\ %
      DP-GP-LVM &&&
        $0.21 \pm 0.07$ & %
        $0.22 \pm 0.03$ & %
        $\phantom{0}78.7 \pm 0.2$ & %
        $117.2 \pm \phantom{1}0.3$ \\ %
      \midrule
      BGP-LVM & \multirow{3}{*}{$50$} & \multirow{3}{*}{$50$} &
        $0.21 \pm 0.08$ & %
        $0.22 \pm 0.04$ & %
        $\phantom{0}11.7 \pm 0.1$ & %
        $\phantom{0}12.9 \pm \phantom{1}0.2$ \\ %
      MRD &&&
        $0.24 \pm 0.07$ & %
        $0.27 \pm 0.04$ & %
        $177.8 \pm 1.3$ & %
        $337.8 \pm \phantom{1}7.9$ \\ %
      DP-GP-LVM &&&
        $0.22 \pm 0.07$ & %
        $0.20 \pm 0.03$ & %
        $\phantom{0}78.0 \pm 0.3$ & %
        $118.5 \pm \phantom{1}2.2$ \\ %
      \bottomrule
    \end{tabular}
  }%
  \caption{Mean squared error between predicted means and ground truth, and runtimes for the missing data experiments.}
  \label{tab:missing_data_results}
\end{table}
The single group model and the DP-GP-LVM perform equally well on this dataset while the MRD model is worse. While it is not surprising that the latter performs poorly due to its failure to extract the underlying dependency structures, the result of the single group model might seem unexpected. However, the BGP-LVM can circumvent the misspecification in the groups by learning a less smooth latent space, reducing dependencies, using a shorter length-scale in the covariance function. An example of this behaviour can be seen in Figure~\ref{fig:pose_track_results_2} comparing \subref{fig:bayes_gplvm_latent01_2} and \subref{fig:gp_dp_latent01_2} we see that the former is less smooth and less certain when moving away from the training data. 
The table also shows that our method is more computationally efficient than the MRD model due to the reduction in free parameters.

\section{Conclusion} \label{sec:conclusion}
We  presented a non-parametric latent variable model with the ability to learn dependency structures in multivariate data. Our approach is capable of organising the observed dimensions into groups that covary in a consistent manner. The model extends previous non-parametric formulations of Inter-Battery Factor Analysis by disentangling the factorisation of the latent space with the characteristics of the generative mapping.

\paragraph{Future Work} We intend to investigate further kernel combinations and latent priors for a wide range of applications. In addition, we intend to adapt the inference procedure to improve scalability and allow for online inference where the number of groupings continually evolves with more data.

%
%
%
%

%
%
%
\section*{References}

\bibliography{ref}

%

\onecolumn
\appendix

\section{Priors on the Latent Variables $\bX$} \label{sec:latent_prior}
In~\S~\ref{sec:model} we allocated an uninformative prior over the latent representation $\bX$ in~\eqref{eqn:model_factor_x}. We note, however, that the prior only appears in the KL term in~\eqref{eqn:fd_kl_x}. This allows for the possibility of using a variety of more informative priors for dealing with specific datasets. As noted by~\cite{bayesian_gplvm}, it is straight forward to extend this model to include a dynamical prior for sequence data.

\paragraph{Dynamic Prior}
Motion capture and pose tracking datasets include time information. The GP-DP model can be extended to include this additional information. The only necessary change is to the prior on the latent variables $\mathbf{X}$. We can define each latent dimension as a temporal latent function drawn from a GP. Therefore,
\begin{align}
\mathbf{x}_q(\mathbf{t}) &\sim \mathcal{GP}(\mathbf{0},k_{\mathrm{x}}(\mathbf{t},\mathbf{t}^{\prime}))\\
p(\mathbf{X} \vert \mathbf{t}) &= \prod_{q=1}^{Q}{\mathcal{N}(\mathbf{x}_{q} \vert \mathbf{0},\mathbf{K}_{\mathrm{x}})},
\end{align}
where $\mathbf{K}_{\mathrm{x}}=k_{\mathrm{x}}(\mathbf{t},\mathbf{t})$.
The variational distribution $q(\mathbf{X})$ is then defined as
\begin{equation}
  q(\mathbf{X}) = \prod_{q=1}^{Q} \mathcal{N}(\mathbf{x}_q \vert \bm{\mu}_q, \mathbf{\Sigma}_q),
\end{equation}
where $\mathbf{x}_q$ and $\bm{\mu}_q$ and $N$-length vectors and $\mathbf{\Sigma}_q$ is a full $N \times N$ matrix.

\section{Stable Calculation of the GP Lower Bound} \label{sec:stable_calculation}
As noted by~\cite{bayesian_gplvm}, care must be taken when calculating the free energy term in~\eqref{eqn:f_d} in~\S~\ref{sec:gp-dp_learning}. We provide a derivation for stable calculation of the free energy.
\renewcommand{\Kuu}{\bm{K}_{\mathrm{uu}}}
\begin{align}
\mathcal{F}_d &= \log \left[ \frac{\beta^{N/2} |\Kuu|^{1/2} }{(2\pi)^{N/2} |\beta\bm{\Psi}_2 + \Kuu|^{1/2}} \exp \left[ -\frac{1}{2} \bm{y}_d^{T} W \bm{y}_d \right] \right] - \frac{1}{2} \beta \psi_0 + \frac{1}{2} \beta \mathrm{Tr} \left[ \Kuu^{-1} \bm{\Psi}_2 \right] \\[5pt]
&= \frac{N}{2}\log[\beta] + \frac{1}{2}\log|\Kuu| - \frac{N}{2}\log[2\pi] - \frac{1}{2}\log|\beta\bm{\Psi}_2 + \Kuu| -\frac{1}{2} \bm{y}_d^{T} W \bm{y}_d - \frac{1}{2} \beta \psi_0 + \frac{1}{2} \beta \mathrm{Tr} \left[ \Kuu^{-1} \bm{\Psi}_2 \right] 
\end{align}
where $W = \beta I_{N} - \beta^{2} \bm{\Psi}_1 \left( \beta\bm{\Psi}_2 + \Kuu \right)^{-1} \bm{\Psi}_1^{T}$.\\[5pt]
\noindent Let $L_u = \mathrm{chol}\left[ \Kuu \right]$ such that $ \Kuu = L_u L_u^{T}$.%
\begin{align}
\Rightarrow \mathcal{F}_d &= \frac{N}{2}\log[\beta] + \frac{1}{2}\log|L_u L_u^{T}| - \frac{N}{2}\log[2\pi] - \frac{1}{2}\log|\beta\bm{\Psi}_2 + L_u L_u^{T}| \\[5pt]&\;\quad -\frac{1}{2} \bm{y}_d^{T} \left[ \beta I_{N} - \beta^{2} \bm{\Psi}_1 \left( \beta\bm{\Psi}_2 + L_u L_u^{T} \right)^{-1} \bm{\Psi}_1^{T} \right] \bm{y}_d - \frac{1}{2} \beta \psi_0 + \frac{1}{2} \beta \mathrm{Tr} \left[ (L_u L_u^{T})^{-1} \bm{\Psi}_2 \right] \\[10pt]
&= - \frac{N}{2}\log[2\pi] + \frac{N}{2}\log[\beta] + \frac{1}{2}\log|L_u||L_u^{T}|  - \frac{1}{2}\log|L_u||\beta L_u^{-1} \bm{\Psi}_2 L_u^{-T} + I_{M}||L_u^{T}| \\[5pt]&\;\quad -\frac{1}{2} \bm{y}_d^{T} \left[ \beta I_{N} - \beta^{2} \bm{\Psi}_1 \left( L_u (\beta L_u^{-1} \bm{\Psi}_2 L_u^{-T} + I_{M} ) L_u^{T} \right)^{-1} \bm{\Psi}_1^{T} \right] \bm{y}_d - \frac{1}{2} \beta \psi_0 \\[5pt]&\;\quad+ \frac{1}{2} \beta \mathrm{Tr} \left[ L_u^{-1}\bm{\Psi}_2 L_u^{-T} \right] \\[10pt]
&= - \frac{N}{2}\log[2\pi] + \frac{N}{2}\log[\beta] - \frac{1}{2}\log|\beta L_u^{-1} \bm{\Psi}_2 L_u^{-T} + I_{M}| \\[5pt]&\;\quad -\frac{1}{2} \bm{y}_d^{T} \left[ \beta I_{N} - \beta^{2} \bm{\Psi}_1 \left( L_u (\beta L_u^{-1} \bm{\Psi}_2 L_u^{-T} + I_{M} ) L_u^{T} \right)^{-1} \bm{\Psi}_1^{T} \right] \bm{y}_d - \frac{1}{2} \beta \psi_0 \\[5pt]&\;\quad+ \frac{1}{2} \beta \mathrm{Tr} \left[ L_u^{-1}\bm{\Psi}_2 L_u^{-T} \right]%
\end{align}
Let $A = \beta L_u^{-1} \bm{\Psi}_2 L_u^{-T} + I_{M}$ and $L_A = \mathrm{chol}\left[ A \right]$ such that $A = L_A L_A^{T}$.
\begin{align}
\Rightarrow \mathcal{F}_d &= - \frac{N}{2}\log[2\pi] + \frac{N}{2}\log[\beta] - \frac{1}{2}\log|A| - \frac{1}{2} \beta \psi_0 + \frac{1}{2} \beta \mathrm{Tr} \left[ L_u^{-1}\bm{\Psi}_2 L_u^{-T} \right] \\[5pt]
&\;\quad -\frac{1}{2} \bm{y}_d^{T} \left[ \beta I_{N} - \beta^{2} \bm{\Psi}_1 \left( L_u A L_u^{T} \right)^{-1} \bm{\Psi}_1^{T} \right] \bm{y}_d  \\[10pt]
&= - \frac{N}{2}\log[2\pi] + \frac{N}{2}\log[\beta] - \log|L_A| - \frac{1}{2} \beta \psi_0 + \frac{1}{2} \beta \mathrm{Tr} \left[ L_u^{-1}\bm{\Psi}_2 L_u^{-T} \right]  -\frac{1}{2} \beta \bm{y}_d^{T} \bm{y}_d\\[5pt]%
&\;\quad +\frac{1}{2} \bm{y}_d^{T} \left[ \beta^{2} \bm{\Psi}_1 L_u^{-T} L_A^{-T} L_A^{-1} L_u^{-1} \bm{\Psi}_1^{T} \right] \bm{y}_d \\[10pt]
&= - \frac{N}{2}\log[2\pi] + \frac{N}{2}\log[\beta] - \log|L_A| - \frac{1}{2} \beta \psi_0 + \frac{1}{2} \beta \mathrm{Tr} \left[ L_u^{-1}\bm{\Psi}_2 L_u^{-T} \right]  -\frac{1}{2} \beta \bm{y}_d^{T} \bm{y}_d \\[5pt]
&\;\quad+\frac{1}{2} \bm{y}_d^{T} \left[ \beta^{2} C^{T} C \right] \bm{y}_d
\end{align}
where $C = L_A^{-1} L_u^{-1} \bm{\Psi}_1^{T}$.\\[5pt]
\noindent Now $\mathcal{L}_{\mathcal{GP}} = \mathcal{F} + \mathrm{KL}\!\big[ q(\bX) \,\|\, p(\bX) \big]$ where $\mathcal{F} = \sum_{d=1}^{D} \mathcal{F}_d$.
\begin{align}
\Rightarrow \mathcal{F} &= - \frac{ND}{2}\log[2\pi] + \frac{ND}{2}\log[\beta] - D\log|L_A| - \frac{D}{2} \beta \psi_0 + \frac{D}{2} \beta \mathrm{Tr} \left[ L_u^{-1}\bm{\Psi}_2 L_u^{-T} \right] \\[5pt]
	&-\frac{1}{2} \beta \mathrm{Tr}\left[\bY \bY^{T}\right] + \frac{1}{2} \beta^{2} \mathrm{Tr}\left[ \bY^{T} C^{T} C \bY \right]
\end{align}
and
\begin{align}
\mathrm{KL}\!\big[ q(\bX) \,\|\, p(\bX) \big] &= \frac{1}{2}\sum_{n=1}^{N} \mathrm{Tr}\left[ \mu_n \mu_n^{T} + \Sigma_n - \log (\Sigma_n) \right] - \frac{NQ}{2}
\end{align}

\section{Kernel Expectations} \label{sec:kernel_expectations}
We provide details of the kernel expectations used in~\S~\ref{sec:gp-dp_learning}. We assume expectations wrt $\bm{Z}$ have already been taken for $\beta$ and the covariance kernels.
\newcommand{\ba}{\boldsymbol{a}}
\newcommand{\bx}{\boldsymbol{x}}
\newcommand{\bz}{\boldsymbol{x}^{\mathrm{u}}}
\newcommand{\bmu}{\boldsymbol{\mu}}
\newcommand{\bS}{\boldsymbol{\Sigma}}
\newcommand{\bPsi}{\boldsymbol{\Psi}}
\begin{align}
\kappa(\bx_i, \bx_j) =&\; \sigma^{2} \exp\!\bigg( -\frac{1}{2} \sum_{q=1}^{Q} \gamma_q \big\| x_{i,q} - x_{j,q} \big\|^{2} \bigg) \\
\psi_0 =&\; \sum_{i=1}^{N} \int \kappa(\bx_i, \bx_i) \, \mathcal{N}(\bx_i | \bmu_i, \bS_i) d\bx_i 
\\
\big[\bPsi_1\big]_{i,j} =&\; \int \kappa(\bx_i, \bz_j) \, \mathcal{N}(\bx_i | \bmu_i, \bS_i) d\bx_i \\
 =&\; \sigma^{2} \prod_{q=1}^{Q} \frac{ 1 }{ (\gamma_q \Sigma_{i,q} + 1)^{1/2} }
            \exp\left[ \frac{ - \gamma_q (\mu_{i,q} - x^{\mathrm{u}}_{j,q})^2 }{ 2 (\gamma_q \Sigma_{i,q} + 1) } \right] \\
        =&\; \sigma^{2}  \frac{ 1 }{ \prod_{q=1}^{Q} (\gamma_q \Sigma_{i,q} + 1)^{1/2} }
            \exp\left[ - \sum_{q=1}^{Q} \frac{ \gamma_q (\mu_{i,q} - x^{\mathrm{u}}_{j,q})^2 }{ 2 (\gamma_q \Sigma_{i,q} + 1) } \right] 
\\
\log\!\Big(\big[\bPsi_1\big]_{i,j}\Big) =&\; \log(\sigma^{2}) - \frac{1}{2} \sum_{q=1}^{Q} \log\!\big(\gamma_q \Sigma_{i,q} + 1\big)
            - \sum_{q=1}^{Q} \frac{ \gamma_q (\mu_{i,q} - x^{\mathrm{u}}_{j,q})^2 }{ 2 (\gamma_q \Sigma_{i,q} + 1) } 
\end{align}
\begin{align}
\big[\bPsi_2\big]_{j,j'} =&\; \sum_{i=1}^{N} \; \int \kappa(\bx_i, \bz_j) \, \kappa(\bz_{j'}, \bx_i) \, \mathcal{N}(\bx_i | \bmu_i, \bS_i) d\bx_i 
\end{align}
\begin{align}
\log\!\Big(\big[\bPsi_2\big]_{j,j'}\Big) =&\; \sum_{i=1}^{N} \; 2\log(\sigma^{2}) - \frac{1}{2} \sum_{q=1}^{Q} \log\!\big(2\gamma_q \Sigma_{i,q} + 1\big)
	-\frac{1}{4} \sum_{q=1}^{Q} \gamma_q \left( x^{\mathrm{u}}_{j,q} - x^{\mathrm{u}}_{j',q} \right)^{2} \\
&\; \quad\;\;  - \sum_{q=1}^{Q} \frac{ \gamma_q }{ 2 \gamma_q \Sigma_{i,q}+1 } \left(\mu_{i,q}
	 - \frac{1}{2} (x^{\mathrm{u}}_{j,q} + x^{\mathrm{u}}_{j',q})\right)^{2} 
\end{align}

\section{Details of the DP Lower Bound} \label{sec:details_dp_bound}
We provide further details about the expressions in the DP lower bound of~\eqref{eqn:dp_bound}.
\begin{align}
  \mathbb{E}_q[\log p(\mathbf{z}_d \vert \mathbf{V})] &= \!\begin{aligned}[t]
    &\sum_{t=1}^{T-1} \bigg[ \phi_{d,t} \left( \Psi{(a_t)} - \Psi{(a_t + b_t)} \right) + \sum_{t'=t+1}^{T} \phi_{d,t'} \left( \Psi{(b_t)} - \Psi{(a_t + b_t)} \right) \bigg]
  \end{aligned} \\
  \mathbb{E}_q[\log p(\mathbf{V} \vert \sigma^{2})] &= \!\begin{aligned}[t]
    &(T-1)\left[\Psi{(w_1)}-\log{(w_2)}\right] + \sum_{t=1}^{T-1} \left[ \left(\frac{w_1}{w_2}-1\right) \bigg(\Psi{(b_t)} - \Psi{(a_t + b_t)}\bigg) \right]
  \end{aligned} \\
  \mathbb{E}_q[\log p(\sigma^{2})] &= \!\begin{aligned}[t]
    &-\log{\Gamma{(s_1)}} + s_1\log{(s_2)} - s_2 \left(\frac{w_1}{w_2}\right) +(s_1-1)\left[\Psi{(w_1)} - \log{(w_2)}\right]
  \end{aligned} \\
  \mathbb{H}[q(\mathbf{V})] &= \!\begin{aligned}[t]
    &\sum_{t=1}^{T} \log{B(a_t,b_t)} - (a_t-1)\Psi{(a_t)} - (b_t-1)\Psi{(b_t)} + (a_t + b_t - 2)\Psi{(a_t + b_t)}
  \end{aligned} \\
  \mathbb{H}[q(\mathbf{Z})] &= - \sum_{d=1}^{D} \sum_{t=1}^{T} \phi_{d,t} \log{(\phi_{d,t})} \\
  \mathbb{H}[q(\sigma^{2})] &= \!\begin{aligned}[t]
    &w_1 - \log{(w_2)} + \log{\Gamma{(w_1)}} + (1-w_1)\Psi{(w_1)}
  \end{aligned}
\end{align}
We have used $\Psi{(\cdot)}$ as the digamma function and $B{(\cdot,\cdot)}$ as the beta function.

\end{document}